\documentclass[letterpaper]{article}
\usepackage[preprint]{aaai2027}
\usepackage[hyphens]{url}
\usepackage{graphicx}
\usepackage{natbib}
\usepackage{caption}
\usepackage{booktabs}
\usepackage{amsmath}   
\usepackage{array}
\usepackage{rotating,fancyvrb,xcolor}
\definecolor{codeframe}{gray}{0.55}
\IfFileExists{fvextra.sty}{\usepackage{fvextra}}{}
\graphicspath{{../../draft/figs/}{./}}
\title{Disclosure-Gated User Simulation for Companion-Agent Evaluation}
\author{Yao Liu, Yu He, Jihao Huang, Xinyang Liu, Feng Zhou}
\affiliations{liuyao14@tsinghua.org.cn}
\begin{document}
\maketitle
\begin{abstract}
Playing the user with a large language model is standard in scalable evaluation, and has a repeatedly diagnosed failure: the simulated user is excessively cooperative, so a system under test can score by the number of questions it asks rather than by making the user willing to speak. We answer with a disclosure gate conditioning information release on the companion agent's behaviour: its state is a ladder of five ordered gates, merged onto three observable depth layers. We specify, ablate, and audit it. The gate is a load-bearing component of the environment: on a published companion-agent benchmark's English corpus, withdrawing per-example gate statements from training displaces the ranking of the 12 systems beyond the noise band of a seed re-run, while per-system scores show no detectable change. We state two acceptance criteria: a ranking must be order-preserving, and absolute scores must be scale-stable. Only one candidate passes both --- the simulator we release --- correlating at 0.993 with the published leaderboard. Prompting a frontier model instead barely moves the ranking while shifting every score up, invisible to ranking-only checks; we supply the specification, ablations, and downstream sensitivity analysis the benchmark's environment lacked.
\end{abstract}
\setlength{\emergencystretch}{3em}
\newcommand{\apxsection}[2]{\par\addvspace{12pt}%
  {\centering\normalfont\Large\bfseries Appendix #1\quad #2\par}%
  \nopagebreak\vspace{4pt}\nopagebreak\noindent\ignorespaces}
\renewcommand{\topfraction}{0.95}
\renewcommand{\bottomfraction}{0.95}
\renewcommand{\textfraction}{0.05}
\renewcommand{\floatpagefraction}{0.85}

\section{Introduction}

Using a large language model to play the user side is common practice, and the verdict on it is negative. Simulated users are excessively cooperative (an ``easy mode'' in which the simulated user volunteers details before being asked, lifting agent success rates above the human baseline \citep{sim2real}), while real users tend to withhold details until prompted \citep{ppol}, so an evaluation cannot separate asking many questions from making the user willing to speak. Mechanisms addressing this failure have precedents \citep{miconsistent, aisp, medconceal, patientact}, but none is specified precisely enough for a third party to rebuild, and none has been measured as an ablatable variable inside an evaluation environment (\S{}2).

We answer with a \textbf{disclosure gate}: information release becomes a function of the companion agent's behaviour. The state is a ladder of five ordered gates merged onto three observable depth layers, each turn's transition determined by the class of that behaviour, with asymmetric fall-back when the relationship is damaged; part of the specification is compiled per example, the rest left for training to internalise (\S{}3, \S{}4). This makes the evaluation ablatable (the gate information removable from training data or runtime prompt on its own, everything else identical item by item), and the mechanism yields a family of readings the rubric total cannot recover (\S{}7).

Evidence is gathered at two layers, the intrinsic and the downstream: conformance of the simulator's own gating, and change in a published leaderboard once the simulator is swapped in (\S{}6). The gate is load-bearing: withdrawing the per-example gate information displaces the ranking beyond the noise band while per-system scores show no detectable change (\S{}6.4). The gating behaviour comes from the synthetic branch of the corpus (withdraw it and the behaviour almost entirely disappears), while a prompted frontier model (\texttt{gpt-5.3}) sustains it only by reading the gate information at runtime (\S{}7). Comparable readings require order-preserving and scale-stable environments; prompting another frontier model (\texttt{gpt-5.1}) as the environment barely moves the ranking, yet its scores rise in concert, invisible to ranking-only checks.

The simulator specified here is the evaluation environment of a published benchmark, CompanionBench \citep{companionbench}, which describes the mechanism in about four hundred words without specification, ablation, human study, negative control, or downstream sensitivity analysis; this paper adds all five and carries over CompanionBench's annotated corpus and displacement baseline (Appendix A). Four contributions follow: \textbf{an executable gate specification with its design space} (\S{}3, Appendices B, G, H); \textbf{validity checks on the instrument itself} --- negative controls for two alternative explanations, with human discrimination and blind paired preference as anchors (\S{}6.1, \S{}6.2, Appendices C--E); \textbf{a test of the gate on out-of-sample human conversations} (\S{}6.2); and \textbf{four manipulations and one downstream measurement} --- data branch, gate information at training and inference, and scale, each altered in turn and read downstream as rank and score displacement (\S{}6.3, \S{}6.4, Appendix D).

\section{Related Work}

\textbf{The field's verdict is negative} on using a large language model to play the human side: the best of 31 simulators reaches a User-Sim Index of 76.0 $\pm$ 1.5 against humans' 92.7 $\pm$ 1.1 \citep{sim2real}, and swapping the simulator moves the same strong assistant's score by 17.2 percentage points \citep{userlm} and a SUT's success rate by up to 9 points \citep{lostinsim}. None of this evidence is from the emotional-companionship domain.

\textbf{Every component of this gating mechanism has precedent.} Remedies act on the behaviour distribution \citep{ppol}, the gap measure \citep{sim2real}, or the training vehicle \citep{userlm}; state-structure remedies are already at the mechanism layer --- controlled state transitions \citep{miconsistent}, hidden-information discovery rates \citep{aisp}, Disclosure Pacing \citep{patientact}, trained latent states \citep{humanlm}. Whether distribution-layer interventions compensate for mechanism-layer failure has no direct test (the nearest partial test leaves a residual of \textit{d} = 0.34 \citep{walkaway}). The simulator-as-instrument lineage is established \citep{balogzhai} --- task-oriented information gating \citep{schatzmann2009}, emotional state used to probe but never to withhold \citep{emous}, ranking reproduction as acceptance criterion \citep{zhangbalog2020, simulatorarena} distinct from behavioural fidelity \citep{simevalir}, whence our two measurement layers; we gate disclosure depth in a way driven by relational behaviour, with retreat after a rupture (\S{}3.2, \S{}3.3).

\textbf{Prior ablations and the closest works stop short of the mechanism itself.} Existing ablations swap the whole simulator \citep{pare} or remove a prompt-side component \citep{simulatorarena}; ours removes a mechanism trained into the weights, gauged against a re-run under a new seed \citep{variance} and judged on order-preserving and scale-stable criteria (\S{}6.4). What is new is the decision rule for the human anchor, not the anchor itself \citep{jonesbergen, ppol}: two-alternative forced-choice with a unidirectional invalidation rule (discrimination can invalidate an indistinguishability claim, not validate it; \S{}6.2, Appendix E). Closest are three: CompanionBench \citep{companionbench} supplies no specification or ablation and our displacement is read against its readings, so its limitations carry over; \citet{avp}'s virtual patient has no behaviour-triggered fall-back and tests load-bearingness within the simulator, whereas we ask about internalisation into the weights and downstream displacement; and \citet{patientact}'s gate is recomputed each turn so re-locking is entailed --- the vehicle and the re-locking trigger are distinct here: a named operation there \citep{douself}, a rupture event here.

\section{Specification of the Disclosure Gate}
\begin{figure*}[t]
\centering
\includegraphics[width=0.79\textwidth]{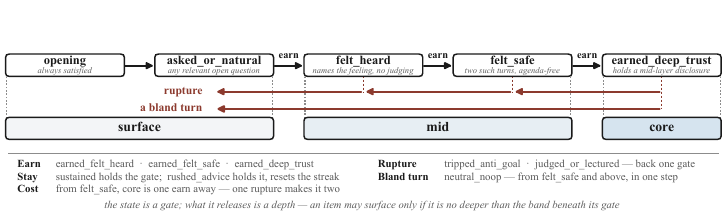}
\caption{Fig 1 --- The disclosure gate: five ordered gates, three released depths, and the eight behaviour classes}
\end{figure*}

The disclosure gate has so far only been asserted (the deeper material must be earned by the companion agent \citep{companionbench}) without state space, transition function, decision criteria, retreat dynamics, per-example compilation, or field-level visibility; this section supplies them at a granularity sufficient for a third party to rebuild the same gate on a corpus of their own.

\subsection{Disclosure Inventory and Gate Legend}

\textbf{The disclosure inventory}. Every persona carries one, listing what that user might say, each item holding four mechanism fields and one switch: \texttt{content}; \texttt{depth}, ordered \texttt{surface} (facts that orient and probe) < \texttt{mid} (feelings) < \texttt{core} (deep vulnerability); \texttt{gate}, which of the five gates the item is mounted behind (\S{}3.2); \texttt{guard}, which guard the item is spoken through while that gate is closed (\S{}3.3); and \texttt{may\_never\_surface}. Material in an unlocked layer is eligible --- a permission, not an obligation. Appendix B.6 gives one persona in full and is the authoritative statement of the mechanism at the level of values.

\textbf{The gate legend} records what the \texttt{gate} and \texttt{guard} values of that persona's items each mean, the meanings coming from two global definition tables (Appendices B.2 and B.3). It is compiled deterministically, carrying no definition the example does not use; the rule set is delivered in two halves (per-item meanings issued with the example, cross-example regularities left for training to internalise; Appendix B.4) and it carries the persona's anti-goals (items that trigger retreat on touch). The simulator thus receives two blocks per example, the inventory's values and the legend's meanings --- the target of the stripping manipulation of \S{}5.2.

\subsection{Gates, State, and the Transition Function}

\textbf{The five gates}, three retaining names from existing work \citep{companionbench}, are ordered as follows, with the three depth layers and the eight behaviour classes set out in Fig 1: $\texttt{opening} < \texttt{asked\_or\_natural} < \texttt{felt\_heard} < \texttt{felt\_safe} < \texttt{earned\_deep\_trust}$.

\textbf{The decision criteria for the five gates} are addressed to judges and annotators, not the state machine, and are cumulative: \texttt{opening} is always satisfied; \texttt{asked\_or\_natural} requires a relevant open question or the topic continuing naturally; \texttt{felt\_heard} requires one accurate reflection or naming, without judging the feeling or overriding it with advice; \texttt{felt\_safe} requires that to be maintained for $\geq$ 2 consecutive exchanges; \texttt{earned\_deep\_trust} requires in addition a \texttt{mid}-layer disclosure held without avoiding it or lecturing. The full wording is in Appendix B.2.

\textbf{State, and the depth it permits.} The gate state is the highest gate earned so far, invisible to the companion agent, held as an index \texttt{level $\in$ \{0,$\ldots$,4\}}; what it permits is a depth: the five gates merge in order into three layers (Fig 1), fixing the deepest layer the turn may reach, with an item eligible exactly when its \texttt{depth} is no deeper than that layer. Five gates are held internally rather than three because merging loses two distinctions (Appendix B.1).

\textbf{The transition function.} Every AI turn is assigned one of eight \texttt{ai\_move} classes, an enumeration shared by synthesis and audit; Table 1 gives the effect of each class on the state.

\begin{table*}[t]
\centering\scriptsize
\setlength{\tabcolsep}{4pt}
\caption{\textbf{Table 1} --- The transition function (effect of the eight behaviour classes; \texttt{fh\_streak} = unbroken run of holding, B.2)}
\begin{tabular}{@{}>{\raggedright\arraybackslash}p{0.232\textwidth}>{\raggedright\arraybackslash}p{0.480\textwidth}>{\raggedright\arraybackslash}p{0.108\textwidth}>{\raggedright\arraybackslash}p{0.093\textwidth}@{}}
\toprule
\texttt{ai\_move} & Effect on \texttt{level} & \texttt{fh\_streak} & Rupture? \\
\midrule
\texttt{earned\_felt\_heard} & \texttt{level $\leftarrow$ max(level, 2)} & +1 & No \\
\texttt{earned\_felt\_safe} & \texttt{level $\leftarrow$ max(level, 3)} if \texttt{level $\geq$ 2}; else holds & +1 & No \\
\texttt{earned\_deep\_trust} & \texttt{level $\leftarrow$ max(level, 4)} if \texttt{level $\geq$ 3}; else holds & Holds & No \\
\texttt{sustained} & Holds & +1 & No \\
\texttt{neutral\_noop} & \texttt{level $\leftarrow$ min(level, 1)} & Holds & No \\
\texttt{rushed\_advice} & Holds & Reset to 0 & No \\
\texttt{tripped\_anti\_goal} & \texttt{level $\leftarrow$ max(0, level $-$ 1)} & Reset to 0 & Yes \\
\texttt{judged\_or\_lectured} & \texttt{level $\leftarrow$ max(0, level $-$ 1)} & Reset to 0 & Yes \\
\bottomrule
\end{tabular}
\end{table*}

The state machine enforces only the ordering; the semantic preconditions of the two highest gates are discharged by the annotator (Appendix B.2) --- if the simulator never produces \texttt{mid}-layer material, no companion behaviour can earn the top gate. One implementation serves both uses: the private director signal at synthesis (\S{}3.4), and \texttt{ai\_move} re-labelling by the audit judge plus scripted replay at evaluation (\S{}5.4). Two properties follow. (i) No gate skipping, enforced by the two conditional \texttt{max} entries: \texttt{core} requires passing \texttt{felt\_heard} $\rightarrow$ \texttt{felt\_safe} $\rightarrow$ \texttt{earned\_deep\_trust} in order, putting social penetration theory's gradualism under state-machine enforcement rather than under a prompt. (ii) From the fourth gate upward, a bland turn costs the state more than tripping an anti-goal, yet triggers no retreat; only the two rupture classes of Table 1 count towards \texttt{retreat\_ok} (Appendix D); the contrast holds at the fourth gate and above only (Appendix B.1); on the human corpus, see \S{}6.2.

\subsection{After Unlocking: Disclosure, Retreat, and Guards}

\textbf{Rules on the disclosure side.} The principal rule is that disclosure is optional (\S{}3.1's may-never-surface switch is its extreme form); otherwise the simulator degrades into a delayed information source, and optionality is why the intrinsic-layer metric takes the difference between conditions rather than the absolute amount disclosed (\S{}5.5). Three further pacing constraints are in Appendix B.3.

\textbf{Asymmetric retreat.} Retreat is triggered by the two behaviour classes that set the rupture flag in Table 1: the state falls back one gate, a guard is placed over the item that was surfacing, and a mild \texttt{withdrawal} is superimposed. The asymmetry carries no coded penalty term: the cost is unequal (at the fourth gate one act of judging drops the state one gate, two to be re-earned before \texttt{core}), and willingness need not return with the state, for the item now carries a guard.

\textbf{Seven guards.} A locked item is spoken through a guard --- six guarding forms and a seventh, \texttt{null}, for none (Appendix B.3). Guards are the observable evidence of retreat, one of the conditions on which \texttt{retreat\_ok} is decided (Appendix D); a bland turn raises no guard, so that observability does not cover it (property (ii) of \S{}3.2).

\subsection{Prompt Assembly and the Visibility Contract}

One shared renderer produces the simulator's system prompt from the persona, the inventory, the legend, and the global statements on pacing and retreat; across synthesis, training, and runtime it differs in two places only: the director signal is present at synthesis and empty thereafter; the anti-goal block does the reverse (Appendix B.5). The training-time and runtime prompts must be byte-identical, both produced by the same function, otherwise no gating behaviour is comparable; the director signal is scaffolding, withdrawn after synthesis, so the hard rule binds the training--runtime pair alone (Appendix B).

\textbf{The visibility contract.} The companion agent does not see the disclosure inventory (otherwise earning depth degrades into asking for the list) but its own behavioural instructions plus three fields, so that \texttt{max\_depth} is earned rather than supplied (Appendix B.5); anti-goals are visible in the trips condition alone (\S{}5.3). The contract is not blind at one point: the audit judge sees the depth labels, so the decision on premature disclosure (\texttt{premature}) is a conformance check, not an independent reconstruction, handled by the permutation test of \S{}6.1.

\textbf{Theory, domain, and the design space.} The mechanism operationalises social penetration theory's penetration / depenetration dynamics \citep{altman-taylor-1973} with rupture--repair superimposed \citep{safran-muran-2000}; ``the gate must be earned'' is anchored in social exchange with disclosure reciprocity \citep{thibaut-kelley-1959,jourard-1971}, and the seven guards in defence mechanisms --- the first three anchors carried over from that work \citep{companionbench}. None of the four components --- an ordered state ladder, a transition predicate, asymmetric retreat, and a per-example gate legend --- is peculiar to companionship (Appendices G and H).

\section{Constructing the Training Corpus}
\begin{figure*}[t]
\centering
\includegraphics[width=0.82\textwidth]{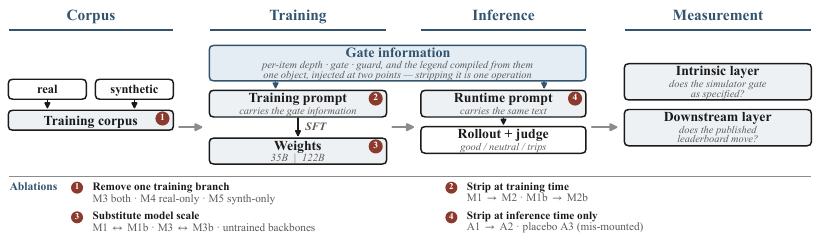}
\caption{Fig 2 --- The pipeline and the four manipulations}
\end{figure*}

The corpus has two branches, real and synthetic. The real branch supplies linguistic texture and genuine reactions: of about 35,000 production conversations \citep{companionbench}, about 31,000 are rated. The synthetic branch supplies the gate itself: whether a deep item surfaces depends on the companion agent's behaviour, with items withdrawn after a rupture, so it must contain, per persona, trajectories on both sides of the gate. The Chinese variants of \S{}6.3 differ only in the branch they were trained on; Fig 2 lays out the pipeline and where the four manipulations (\S{}1) cut.

\subsection{The Two Branches}

The real branch alone cannot carry the gate --- three gaps and one difference of principle. The companion agent is rated good in only 2.28\% of rated conversations (0.00\% and 0.12\% where the gate matters most); only 2.18\% of rated users have both a good- and a poor-rated conversation; and the gate is per turn while the real branch is annotated per item (Appendix F). The difference of principle is direction: in the real branch the conversation comes first and the gate is inferred; in the synthetic the inventory comes first and the gate imposed --- only the synthetic branch contains a genuine manipulation. Each branch has its use --- the synthetic raises gating ability, the real distributional fidelity (\S{}6.3).

\subsection{How the Branches Are Built}

Three exclusions (Appendix F) leave the real branch at about 13,000 conversations and 286,000 trainable turns; its gate is annotated by disposition rather than transcript --- how defended the item is, not when it surfaced (Appendix F).

For the synthetic branch, seeds from the same annotations are arranged into cells across 17 scenarios $\times$ 4 attachment types \citep{companionbench} (Appendix F). Trajectories hold the persona fixed and vary only the companion agent, each following one of six behavioural scripts --- three constant, three arcs (good-to-bad, bad-to-good, good-bad-repair); a gate earned then violated appears only within one conversation. Each turn, a director classifies the reply into one of the eight behaviour classes and the state machine updates the gate; which layer the turn may reach is decided by the state machine, not any model. The English corpus is a downstream transcreation of the Chinese trajectories, paired item by item.

\subsection{The Training Example and the Corpus Sizes}

In training the roles are inverted (the simulated user's turns in the \texttt{assistant} position, the only loss-bearing position) and a shared renderer makes the training prompt byte-identical to the runtime one (\S{}3.4); what stripping removes, gate legend semantics included, is listed in \S{}5.2 (Appendix J). The Chinese set is about 33,000 conversations and 714,000 trainable turns (synthetic to real 60:40 \citep{companionbench}), the English to about 22,000 and 454,000 (Appendix F; field visibility in Appendix B).

\section{Experimental Setup}

\subsection{Research Questions}

Four questions organise the experiments, one per subsection of \S{}6: whether the gate is measured, whether it holds for real people, where the gating comes from, and whether the conclusion changes downstream --- the first three at the intrinsic layer, the fourth downstream; the two layers can give different answers, and neither substitutes for the other (\S{}7). Every reading except the human studies of \S{}6.2 is produced with an LLM judge; perplexity and the Wasserstein distances are computed directly.

\subsection{Model Variants}

The candidate set is ordered by mechanism manipulation, not model name: eight trained variants (M1--M5 plus 122B counterparts M1b, M2b, M3b), three runtime configurations on the same weights (A1 intact, A2 stripped, A3 mis-mounted), two 235B reference simulators, untrained backbones at three scales, and five external models --- every label defined row by row in Appendix I. The release candidates are M1b and M3b (Appendix K).

Stripping removes the same objects whether applied at training time or at inference: the gate-semantics and guard-semantics sections and the per-item mounting, keeping the material, the anti-goals, and the static template of the disclosure policy (Appendix J). A1 labels M1 under the inference-time manipulation (the same weights), so both ablations start from the same intact model; A3's mis-mounting is the placebo --- the annotation stays in place and only the material is scrambled within a persona, giving wrong mounting relations with identical shape (Fig 2).

\subsection{Corpora and Protocol}

Six corpora are used --- an intrinsic-layer persona set (100 personas per language $\times$ 3 conditions), an English extension set disjoint from it (321 personas), a real-conversation set (999 sessions, with 600 simulator re-runs on the same personas under a consistent protocol), a real held-out set (500), the human-study materials, and the downstream corpus (12 SUTs $\times$ 100 personas $\times$ 7 environments = 8,400 sessions) --- tabulated in Appendix F. The companion agent runs under one of three condition blocks --- good, neutral, trips --- differing only in the anti-goal placeholder (\S{}3.4); the contrast and the veto take the good and trips blocks, and every rollout decodes at temperature 0.7.

The real-conversation set is stratified to exclude, by construction, sessions with no deep material to disclose, so real people's depth in \S{}6.2 is not an unbiased estimate. The downstream corpus is one run of the same SUTs in each of seven evaluation environments (an environment being the user simulator conversing with the SUT): the frozen reference, its re-run under a new seed (setting the noise floor and the noise band), and five candidates (Table D5, Appendix D); the SUTs span that leaderboard's capability range \citep{companionbench}. The three human-participant studies, all in \S{}6.2, use no LLM judge (verbatim materials in Appendix E). The real corpus is Chinese only, so \S{}6.3's data ablation and perplexity are Chinese alone and the rest chiefly in English (Limitations).

\subsection{Judges}

The main judge is \texttt{deepseek-v4-pro} (non-reasoning mode), a single judge carrying both the turn-by-turn gate audit and the rubric scoring: it is from a different family from the backbones of all mechanism variants, is the same judge as the published leaderboard we compare against \citep{companionbench}, and serves as the external reference for distributional fidelity (Appendix I); the body cites its own readings only in \S{}6.1, on the side of a gatekeeping failure, which self-preference \citep{playfavorites} would understate rather than exaggerate. The second judge is \texttt{claude-opus-4-8}, re-judging A1, A2, and M2 on the same items and batch, to test whether signs and ordering change with the judge (Appendix C); three limits of this instrument's resolution are in Limitations.

\subsection{Metrics}

The intrinsic-layer verdict rests on one primary endpoint and one veto criterion (the core-reach rate under the trips condition), the downstream layer on five further quantities, all defined in Table D4 (Appendix D). The mid-layer disclosure contrast measures the difference in disclosed depth for one persona under two conditions: the deepest layer reached is reduced to two tiers by collapsing core into mid (\texttt{surface} 0, \texttt{mid} and \texttt{core} both 1), the trips value is subtracted from the good value, and the mean is taken over personas: +1 for a persona reaching \texttt{mid} under good but stopping at \texttt{surface} under trips, $-$1 in reverse (formula in Appendix D). The intrinsic layer is also measured on three further counts, on which \S{}6.3 relies qualitatively: surface fidelity (six rubric dimensions, Appendix C), predictive fit (perplexity), and distributional fidelity (Wasserstein-1, W1); the five secondary gate-audit metrics, the gate-advance rate among them, are defined in Appendix D.

Both quantities have degenerate extremes, stated with their definitions in Appendix D: the mid-layer disclosure contrast scores 0 both for a simulator that discloses unconditionally and for one that discloses nothing --- it rewards only the dependence of depth on the companion agent's behaviour --- while the veto criterion's extreme belongs to the simulator that discloses nothing, so a verdict on it alone would rate the worst simulator best. Human discriminability depends on no automatic instrument (\S{}6.2).

\section{Results}
\begin{figure}[t]
\centering
\includegraphics[width=\linewidth]{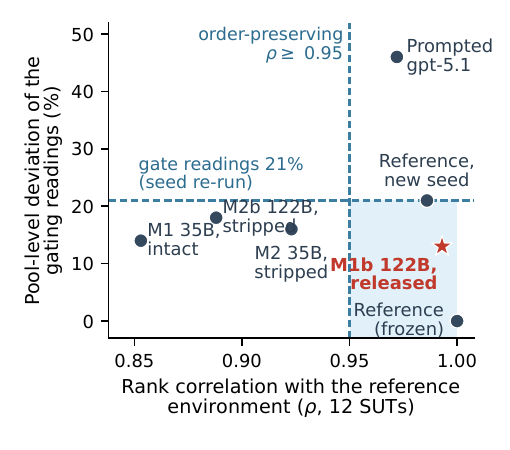}
\caption{Fig 3 --- Rank agreement and gate readings disagree about which environments pass}
\end{figure}

\subsection{Is the Gate Measured?}

Readings come from the intrinsic-layer persona set of \S{}5.3 (Appendix D). The floor is effectively zero: the untrained previous-generation 235B backbone reads +0.090 in Chinese and +0.140 in English on the mid-layer disclosure contrast, with depth barely depending on the companion agent's behaviour (the failure \S{}1 cites), and training lifts it to +0.530 and +0.680. The veto criterion catches something else: under the trips condition \texttt{deepseek-v4-pro} and that backbone still push 19\% and 32\%--37\% of sessions to \texttt{core} against 3\%--8\% for sound gatekeeping; a model can pass the contrast and still fail the veto.

Both alternative explanations are ruled out by their controls: the readings survive a within-persona shuffle of the depth labels (on the untruncated contrast) and a permutation of the condition labels, and a blinded re-run of the judge prompt shifts them by at most 0.06 (Appendix D) --- they vary systematically with the condition, and that variation comes from the gate.

\subsection{Does the Gate Hold for Real People?}

\textbf{The ordering in time holds; the total depth does not.} In the real-conversation set (\S{}5.3) deep material appears more often after a gate has opened, but the gate does not predict how deep a person eventually goes --- one trajectory per session, an observational comparison. A within-session shuffle of the companion agent's actions (same multiset, different opening moments) drives \texttt{premature} below and \texttt{retreat\_ok} above their reference distributions in all six tests, with the annotations unchanged, so the two confounds --- gate-label visibility and the shared judge call --- do not apply (Appendix D).

\textbf{Neither positive association survives control.} With the gate labels visible to the judge, the tercile endpoint (Appendix D) in \texttt{max\_depth} (+0.187, CI [+0.111, +0.260]) and the covariate-adjusted regression of \texttt{max\_depth} on the gate-advance rate (+0.490) are positive; splitting the judge call into two, one per party, with no shared context, drops the coefficient to +0.150 (CI [$-$0.045, +0.345]), and a session-level rating from a separate pipeline gives only +0.018 --- shared method variance accounts for most of the association.

\textbf{People can tell the simulator apart, yet prefer it.} In all three tasks the raters are people, no LLM judge (Appendix E); given a real session's context and two candidate next utterances --- the real person's and the release candidate M3b's --- they take M3b's for the real person's in 37.8\% of judgements (n = 360, CI [31.9\%, 43.5\%], \textit{p} = 4$\times 10^{-6}$). The interval lies wholly below 50\%, so this paper does not claim M3b is indistinguishable from a real person. On 100 persona-matched blind pairs they choose M3b as more like a real person in 68.3\% of judgements (205/300, \textit{p} < 0.0001), arbitrating two LLM judges that had disagreed. And when 5 raters relabel the layer of 145 disclosure items, the deep end cannot be reproduced (Appendix E) --- depth is subjective there.

\subsection{Where Does the Gating Come From?}

\textbf{The synthetic branch buys the gating.} Three 35B variants (M3, M4, M5) differ only in which branch they carry (Chinese only, \S{}5.3); real production conversations do not produce the behaviour. The mid-layer disclosure contrast reads +0.610 synthetic-only, +0.520 synthetic-plus-real, +0.080 real-only; the primary endpoint, the paired difference from real-only to synthetic-only, is $\Delta$ = +0.532, \textit{p} = 0.0001: removing the synthetic branch almost entirely eliminates the gating, leaving a floor-level reading (a confound: step counts differ). From synthetic-plus-real to synthetic-only, $\Delta$ = +0.090, CI [$-$0.040, +0.220] containing 0: real data on top of the synthetic branch adds no detectable improvement. On the rubric's six dimensions the synthetic-only variant is indistinguishable from the mixture on four, behind on one, ahead on one.

\textbf{Gating is in the weights, not the prompt.} With stripping at inference only (A1 to A2), the contrast differs by $-$0.022, CI [$-$0.093, +0.053] (\textit{n} = 321), the effect bounded above in magnitude by 0.093: the gating is internalised into the weights, and the mounting relations seen in training need not be supplied again at runtime. The placebo A3 (reads no higher than stripped; Appendix D) rules out unfamiliar format; M2 never saw the mounting relations and still learned conditional disclosure.

At training time (Table D6, Appendix D) M2 and M2b remain well above the same-generation untrained 35B backbone (+0.240 in both stripped and intact states, so gate information is conferred by training, not by the prompt); the three-tier verdict is undetermined at 35B (paired $\Delta$ = $-$0.080) and equivalent at 122B ($\Delta$ = $-$0.047, CI [$-$0.170, +0.070], at $\varepsilon$ = 0.15): the released scale falls on the side judged equivalent, so conditional disclosure can be learned from the trajectories themselves. The verdict is confined to the intrinsic layer (\S{}6.4), and stripping retains the static policy template --- the claim is only that mounting relations need not be stated explicitly. Re-judging with the second judge leaves all six intervals (three paired comparisons under each judge) containing 0.

\subsection{Does the Conclusion Change Downstream?}

Replacing the instrument does change the conclusion. Three groups are compared --- training-time stripping within pairs, a prompted frontier model, and the four trained variants, each against the reference (Table D5) --- the judge held constant, against an empirical noise band: the frozen reference re-run under a new seed gives $\rho$ = 0.986 (the noise floor), a displacement of 1 position (the noise band), and a top-5 overlap of 5/5. The primary endpoint downstream is the largest rank displacement.

Both criteria read the rubric total on that benchmark \citep{companionbench}: \textbf{order-preserving} --- rank correlation between the two leaderboards for the same 12 SUTs at least 0.95 (observed 0.972 and 0.993, the rest no higher than 0.930); \textbf{scale-stable} --- no detectable change in any SUT's rubric total across the two environments.

Fig 3 plots the seven environments on rank correlation and gating-reading deviation. Stripping changes the ordering without changing the scores: at both scales the largest displacement is 3 positions, three times the noise band; the gate is load-bearing. The prompted frontier model does the reverse: every SUT's rubric total shifts up by 0.521 points (scores not comparable with any historical result) while the ordering barely moves. On the gating readings, the four trained variants stay within the 21\% set by the seed re-run (13\%--18\%); the prompted model reaches 46\%, with \texttt{core\%} and \texttt{deep\_trust\%} up to +93\% and +83\%. Only M1b satisfies both criteria and is released as the evaluation environment.

\section{Discussion}

\textbf{What counts as an acceptable environment, and why a single quantity misleads.} \S{}6.4 states two criteria: an order-preserving ranking and scale-stable scores. Prompting a frontier model is the least visible failure: the ranking differs by 2 positions while the score scale shifts up 0.521 points. Meeting both rests on a mechanism specified, trained into the weights, and auditable: stripped at inference, \texttt{gpt-5.3} loses two fifths of it while M1 shows no detectable change (Appendix D); neither criterion guarantees the interaction resembles a real person \citep{simevalir}, and people can tell the release candidate apart (\S{}6.2). Only M1b passes both. The gate gives a family of readings (premature disclosure among them) the rubric total cannot recover, which repeatedly disagrees with it: the gating readings accept all four trained variants while the ranking separates them, and M1 is the prompted model's mirror image --- scores indistinguishable, rank correlation 0.853. The two layers part company: M1b and M2b, equivalent intrinsically at $\varepsilon$ = 0.15, are displaced downstream by 3 positions --- three times the noise band.

\textbf{Whether the simulator needs a larger model.} That depends on its use \citep{psibench}. As a probe, 35B suffices --- a 3.5$\times$ increase within one generation produces no detectable effect on the gating, though the Chinese side is undetermined (Appendix D). As an evaluation environment, 122B is required: M1b correlates with the reference at 0.993 against 0.853 for M1, the gating readings already saturated at 35B. The two quantities point in opposite directions, and the release follows the rubric total.

\section{Conclusion}

This paper specifies information release as a conditional gate --- state ladder, transition and retreat rules --- checkable turn by turn by replay. The gating behaviour comes from the synthetic branch of the corpus; the gate is load-bearing --- rankings are displaced, per-system scores show no detectable change --- and a prompted frontier model is order-preserving but not scale-stable.

Neither criterion looks at whether the interaction resembles a real person: when a simulator is replaced, the ranking, the absolute scores, and the gate's readings should be reported (\S{}7). For cross-domain transfer we give only domain-independent components and unvalidated design sketches.\label{p2:lastcontent}

\section*{Limitations}

\textbf{Circularity and the resolution of the instrument.} The audit judge can see the depth labels, so the premature-disclosure verdicts are only a conformance check; on the real branch the gate is an inference conditioned on the outcome, not a manipulated variable (\S{}4.1). The transition function's determinism does not make the readings objective. Resolution is limited in three ways: turn-by-turn behaviour annotation agrees only at $\kappa$ = 0.427, so absolute values may not be cited across batches; the primary endpoint carries a detectable judge effect; and the deep end cannot be reproduced (\S{}6.2). Future work: an annotator blind to the depth labels, and a depth quantity that does not rest on labels at the deep end.

\textbf{The gate is a specification, not a descriptive model of disclosure between people.} Disclosure to an AI is subject to a disinhibition effect that runs counter to the ladder and that the specification does not represent; on the real-conversation set the prediction of \S{}6.2 is only half borne out: the total depth does not hold. Turns flagged \texttt{realized\_divergence} on the real branch --- utterances running counter to the persona's disposition --- were meant to be down-weighted in training; the flag never entered the samples, so the down-weighting never took effect. Future work: estimating each item's gate by blind annotation without access to the conversation.

\textbf{Coverage of the evidence.} The downstream evidence is English only, so there is no direct evidence that M3b is order-preserving; the two acceptance criteria were not pre-registered but reconstructed after the total-depth conclusion of \S{}6.2 was withdrawn, and a displacement in either direction indicates only that the gate is load-bearing, not that it does harm. This paper has no evidence as to whether the judge's scores agree with how people actually feel. Readings from the two language versions of each judge prompt are not compared as absolute values (Appendix C). The human anchors rest on the Chinese corpus alone and the 5 raters were all in-house; the runtime ablation's sample size supports only an upper bound (\S{}6.3). Future work: external annotators, and a corpus of English human conversations.

\textbf{Training stops at supervised fine-tuning.} The released simulator underwent no preference-optimisation stage, and one defect may be attributable to that absence: the collapse in single-message length (53.5\% of turns are no more than 8 Chinese characters, against 27.1\% for real people). Future work: preference optimisation on the gating objective, to test whether it repairs that defect without sacrificing the gating.

\section*{Ethics Statement}

The real corpus consists of de-identified, licensed production logs, and its use and redistribution are governed by the three rulings in Appendix K. The text of those conversations is visible only to annotators inside the project and enters no release artifact. Records outside the evaluation's scenario scope, records carrying no disclosure-inventory annotation, and acute-crisis records are dropped in their entirety at the basic filtering stage. The item-level depth annotation and the three human tasks were carried out by the same people inside the project, under the same visibility restriction; the 5 raters are members of the project team, and Appendix E gives the item counts and each rater's workload. What this paper offers is a comparability criterion for an evaluation environment, not a safety certification; a system that scores well under it is not thereby endorsed for deployment.

\bibliography{refs}

\apxsection{A}{Comparison with the Published Version}

The simulator specified in this paper is the evaluation environment used by the published benchmark. That paper describes this simulator, and its training, in a section of about four hundred words, and reports fidelity on two axes (together with a perplexity check); this paper does not re-run that evaluation of the published simulator. Table A1 is organised into two groups, specification and evidence. The middle column reports the result of a line-by-line check against the publicly available arXiv version (v2) of that paper, and the right-hand column indicates where each item is taken up in this paper. The boundary between the two is drawn line by line, so that what does not overlap can be checked directly.

\begin{table*}[t]
\centering\footnotesize
\setlength{\tabcolsep}{4pt}
\caption{\textbf{Table A1} --- The gating mechanism: the published description and what this paper specifies, line by line}
\begin{tabular}{@{}>{\raggedright\arraybackslash}p{0.371\textwidth}>{\raggedright\arraybackslash}p{0.265\textwidth}>{\raggedright\arraybackslash}p{0.295\textwidth}@{}}
\toprule
Item & As published & What this paper specifies (where) \\
\midrule
\textit{Specification} & --- & --- \\
Names of the gates & Five stated but not listed; the figure shows three, one per depth layer & Five, all named (\S{}3.2) \\
Decision criteria and trigger conditions for the five gates & Not specified & \S{}3.2 \\
The transition function and the re-locking of unlocked layers & Repeated opening and closing, and re-locking, mentioned; no transition function given & \S{}3.2 (Table 1), \S{}3.3 \\
A classification of guard language & Not mentioned & Six observable forms plus \texttt{null}, each with its observable marker (\S{}3.3, Appendix B) \\
The compilation algorithm for the gate legend & Not specified & \S{}3.1 (deterministic compilation) \\
The visibility contract: which fields each party sees & Specified: a party-by-field matrix over SUT, simulator and judge & Adds how rendering differs across synthesis, training and runtime (\S{}3.4, Appendix B) \\
A single renderer shared by the training and runtime paths & Not specified & \S{}3.4 \\
The two branches of the training corpus, their construction and scale & About one paragraph of prose & All of \S{}4, Appendix F \\
The rejected design space & Not specified & Appendix H (twelve entries, each with its reason, most with a reopening condition) \\
A domain-general template and cross-domain instantiations & Not specified & Appendix G (three design sketches, none implemented or validated) \\
\textit{Evidence} & --- & --- \\
Whether intrinsic-layer fidelity readings come from the gate itself (negative control) & Not tested; that paper's \S{}7 runs a separate permutation control on its own gating readings & \S{}6.1, Appendix D \\
Whether the gate holds on out-of-sample human conversations & Not tested & \S{}6.2, Appendix F \\
A human anchor & Not tested; that paper reports no human or counsellor baseline & \S{}6.2, Appendix E \\
Judge robustness & Test--retest and pairwise agreement among three cross-family judges & Adds a second, cross-family judge alongside this paper's main judge, and turn-by-turn annotation agreement (\S{}5.4, \S{}6.3, Limitations) \\
Whether the gate is load-bearing: training side, runtime, and scale & Not tested & \S{}6.3, Appendix D \\
Downstream consequences: SUT rankings and scores once the gate is removed & Not tested & \S{}6.4 \\
\bottomrule
\end{tabular}
\end{table*}

Three items of data and experimental material are carried over from that paper (what is carried over at the conceptual level is in \S{}3.4): the annotated corpus on which the real branch of the training corpus rests, the scenario and attachment-type partition, and the leaderboard used here as the downstream displacement baseline (together with its rubric total and its judge). The displacement measured in \S{}6.4 is therefore relative to that paper's published readings, and its limitations are at the same time the limitations of this paper's displacement baseline.

\apxsection{B}{The Gating Rule Set}

\subsection*{B.1 The Disclosure State Machine}

\textbf{The state is the highest gate earned so far}, and the five gates are ordered (\S{}3.2 gives the prose statement; what follows is the formal definition):

\begin{equation*}
\resizebox{\columnwidth}{!}{$\displaystyle \text{opening} < \text{asked\_or\_natural} < \text{felt\_heard} < \text{felt\_safe} < \text{earned\_deep\_trust}$}
\end{equation*}

The state is held as an index $\text{level} \in \{0,\dots,4\}$, \texttt{level = k} meaning the $k$-th gate is earned, initial value 0 (\texttt{opening}, always satisfied); the transition formulae (Table 1) are defined over it.

The five gates merge in order into three observable depth layers, and what results is the deepest layer to which the turn may descend, $\text{allowed\_depth}$:

\begin{equation*}
\resizebox{\columnwidth}{!}{$\displaystyle \text{allowed\_depth}(\text{level}) = \begin{cases} \text{surface} & \text{level} < 2\\ \text{mid} & 2 \le \text{level} < 4\\ \text{core} & \text{level} = 4 \end{cases}$}
\end{equation*}

An inventory item becomes eligible once its \texttt{depth} is no deeper than $\text{allowed\_depth}$; eligibility is permission, not obligation. The state can fall back (B.3), and deep trust is not restored immediately upon repair.

\textbf{The implementation holds five gates internally rather than the three layers directly}, because merging loses two distinctions, each with behavioural consequences. \texttt{felt\_heard} (2) and \texttt{felt\_safe} (3) are both \texttt{mid}, yet only once \texttt{felt\_safe} is earned can an \texttt{earned\_deep\_trust} lift the state to \texttt{core} (the conditional \texttt{max} of Table 1) --- a record holding only ``mid reached'' cannot decide whether that step takes effect. \texttt{opening} (0) and \texttt{asked\_or\_natural} (1) are both \texttt{surface}, yet at \texttt{level = 1} the turn's directive is ``probing permitted'' (\texttt{may\_test}) and at \texttt{level = 0} it is \texttt{hold}: same depth, different licensed actions.

\textbf{No gate skipping.} Reaching \texttt{core} requires passing \texttt{felt\_heard} $\rightarrow$ \texttt{felt\_safe} $\rightarrow$ \texttt{earned\_deep\_trust} in order. A two-turn case from the initial state: \texttt{earned\_deep\_trust} is assigned first (level = 0 < 3, so it does not take effect and the state does not move), then \texttt{earned\_felt\_heard}, lifting the state to 2. The final state is level = 2, not 4.

That holds only when the state starts from 0: had it been 3, the same pair would have lifted it to 4 and then held it at 4. The no-gate-skipping rule constrains each single step's landing point; it does not make a sequence's final state independent of order --- the case above is itself an instance of order sensitivity, on which \S{}3.2, property (i), rests.

\textbf{A level-by-level comparison of the bland turn against rupture.} Property (ii) in the body states that from the fourth gate upward, a bland turn costs the state more than tripping an anti-goal. Table B1 gives every starting point, and the comparison holds only at the higher gates:

\begin{table*}[t]
\centering\footnotesize
\setlength{\tabcolsep}{4pt}
\caption{\textbf{Table B1} --- The bland turn against rupture, level by level (by starting level)}
\begin{tabular}{@{}>{\raggedright\arraybackslash}p{0.261\textwidth}>{\raggedright\arraybackslash}p{0.248\textwidth}>{\raggedright\arraybackslash}p{0.261\textwidth}>{\raggedright\arraybackslash}p{0.144\textwidth}@{}}
\toprule
Starting level & \texttt{neutral\_noop} $\rightarrow$ & \texttt{tripped\_anti\_goal} / \texttt{judged\_or\_lectured} $\rightarrow$ & Which costs more \\
\midrule
4\texttt{earned\_deep\_trust} & 1\texttt{asked\_or\_natural} & 3\texttt{felt\_safe} & The bland turn \\
3\texttt{felt\_safe} & 1\texttt{asked\_or\_natural} & 2\texttt{felt\_heard} & The bland turn \\
2\texttt{felt\_heard} & 1\texttt{asked\_or\_natural} & 1\texttt{asked\_or\_natural} & Equal \\
1\texttt{asked\_or\_natural} & 1\texttt{asked\_or\_natural} (holds) & 0\texttt{opening} & \textbf{Rupture} \\
0\texttt{opening} & 0\texttt{opening} (holds) & 0\texttt{opening} (floor) & Equal \\
\bottomrule
\end{tabular}
\end{table*}

The cause is a difference of shape: applied repeatedly, $\min(\text{level},1)$ stops at 1 whereas $\max(0,\text{level}-1)$ continues to 0, so the current rule penalises the first bland turn more heavily and consecutive ones more leniently. The behaviour at the lower gates is a defect in this specification: at \texttt{asked\_or\_natural} and below a bland turn has no consequence at all while tripping an anti-goal still costs a gate --- the reverse of the design intent. It is a by-product of $\min(\text{level},1)$ stopping at 1 there, not deliberate, and should be corrected in a later version. Replay readings for four rules (the current and three alternatives) are in Appendix D.4.

Three anchors are carried over from the published benchmark, the fourth added here: social penetration theory for disclosure depth (four stages merged into three observable layers, gradualism enforced by the conditional \texttt{max}, depenetration realised as retreat); rupture--repair for relational events; social exchange with disclosure reciprocity for ``the gate must be earned'' (only the half conditioning release on the other party's behaviour, not the symmetric return); and defence mechanisms for the seven guards (their surface linguistic form, not their psychodynamic interpretation).

\textbf{One place the theory does not cover, where this paper makes its own choice.} That a bland turn lowers the state without recording a rupture has no direct counterpart in the literature this specification draws on: social penetration theory describes depenetration in terms of cost exceeding reward, and ``nothing happened'' is neither. The choice made here is that a bland turn does no relational damage (no rupture recorded, no guard raised) but neither maintains what has been unlocked --- otherwise a long session of uniformly perfunctory replies would stay unlocked at \texttt{core} on the grounds that no anti-goal was ever touched.

\subsection*{B.2 Two Wordings of Each Gate, and \texttt{fh\_streak}}

The five gates' decision criteria (third-person and decidable: ``what the other party has done'') are given in \S{}3.2, and judges and annotators use them to decide whether a gate has opened. Three things they do not cover follow.

\textbf{First, the criterion for ``accurately reflected''.} The reply leaves the user feeling it genuinely understood what was just said, rather than offering a stock phrase, a change of subject, or an answer to a different question.

\textbf{Second, each gate has two wordings.} \S{}3.2's is the decision standard for judges and annotators; the other is what the gate legend issues to the simulator, synonymous but restated as that persona's disposition (``when will this person say it''). The judge evaluates the companion agent's behaviour, the simulator re-enacts a person's reaction; that the wordings differ is precisely how the gate shows itself to be a stimulus--response contract. Table B2 is the compiled version (B.6's worked example is drawn from it):

\begin{table*}[t]
\centering\footnotesize
\setlength{\tabcolsep}{4pt}
\caption{\textbf{Table B2} --- The wording of the five gates in the gate legend (the disposition wording on the simulator side)}
\begin{tabular}{@{}>{\raggedright\arraybackslash}p{0.410\textwidth}>{\raggedright\arraybackslash}p{0.538\textwidth}@{}}
\toprule
Gate & Wording in the gate legend \\
\midrule
\texttt{opening} & comes out right in the opening \\
\texttt{asked\_or\_natural} & flows once asked, or once the topic naturally lands there \\
\texttt{felt\_heard} & only admitted once you feel accurately heard and not judged \\
\texttt{felt\_safe} & only once a real sense of safety is established \\
\texttt{earned\_deep\_trust} & may only surface once the AI has truly held the relationship steady \\
\bottomrule
\end{tabular}
\end{table*}

\textbf{Third, the division of labour between \texttt{fh\_streak} and \texttt{felt\_safe}.} The state carries a counter \texttt{fh\_streak} of the turns in an unbroken run of holding: \texttt{earned\_felt\_heard}, \texttt{earned\_felt\_safe}, and \texttt{sustained} each add +1; \texttt{rushed\_advice}, \texttt{tripped\_anti\_goal}, and \texttt{judged\_or\_lectured} reset it to 0; \texttt{earned\_deep\_trust} and \texttt{neutral\_noop} leave it unchanged (Table 1). It takes no part in any transition (no branch in Table 1 reads it to decide \texttt{level}): the \texttt{felt\_safe} standard's ``$\geq$ 2 consecutive exchanges'' calls for a semantic judgement --- what counts as ``an exchange'', and whether an agenda was pushed, are not deterministically executable predicates --- and is discharged by the annotator when emitting \texttt{earned\_felt\_safe}, the state machine enforcing orderedness alone. \texttt{fh\_streak} therefore serves auditability: one can check afterwards that each \texttt{earned\_felt\_safe} is preceded by an unbroken run of holding.

\subsection*{B.3 Disclosure, Retreat, and Guards}

\textbf{Guards are what make retreat observable.} The gate state itself is invisible to the other party, so if a retreat changed only that hidden index and not the way the user speaks, there would be no way to check that it had happened; the guard placed over a surfacing item is that observable change (\S{}3.3).

\textbf{Three further constraints govern how the disclosure is played} (optionality is in \S{}3.3):

\begin{enumerate}
\item No dumping: even when several items are eligible at once, they are spread across turns and drawn out by the other party's responses.
\item Probe first: before disclosing \texttt{mid} or \texttt{core} material, the simulator usually tests the water first (half a sentence, or a guarded version), and states it in full only once the other party has responded.
\item Hold the \texttt{voice}: the persona's register is maintained throughout, and intensity is lowered in line with dispositions such as ``tends to play things down''.
\end{enumerate}

\textbf{Retreat and re-locking.} The trigger and the three-part reaction are in \S{}3.3; two further details follow.

\textbf{The specification and the implementation are not strictly equivalent here.} The transition function descends one gate (\texttt{max(0, level $-$ 1)}), whereas the retreat rule issued to the simulator is stated in depth layers (``\texttt{unlocked\_depth} falls one layer''). Falling from \texttt{felt\_safe}(3) to \texttt{felt\_heard}(2) stays within \texttt{mid}, so the layer does not change, while a rule stated in layers would be read as ``fall back to \texttt{surface}''. Every reading here is computed by the transition function at gate level, so no reported quantity is affected; what is affected is only the simulator's understanding of ``how far to fall'' --- one concrete source of the self-limitation: re-locking is a specification-level requirement, and the instrument does not verify that it is stably enforced.

\textbf{The cost of rebuilding is not explicitly coded.} The implementation holds neither a repair timer nor a penalty term; the cost follows entirely from one asymmetry in Table 1: lifting a gate has a precondition (the conditional \texttt{max}), lowering one does not. Returning to \texttt{earned\_deep\_trust} after a fall-back from \texttt{felt\_safe} requires two earning moves again, each adjudicated separately; a single repair cannot discharge both. The surface form of each of the seven guards is in Table B3.

\begin{table*}[t]
\centering\footnotesize
\setlength{\tabcolsep}{4pt}
\caption{\textbf{Table B3} --- The surface form of the seven guards}
\begin{tabular}{@{}>{\raggedright\arraybackslash}p{0.458\textwidth}>{\raggedright\arraybackslash}p{0.490\textwidth}@{}}
\toprule
Guard & Surface form \\
\midrule
\texttt{minimized} & plays the weight down (``it's not that bad'', ``it wasn't serious'') \\
\texttt{rationalization} & covers the feeling with a reasonable explanation (``I do see that she has her own difficulties'') \\
\texttt{intellectualization} & abstract, analytic register, with no feeling words \\
\texttt{deflection} & changes the subject, or turns the question back \\
\texttt{humor} & laughs it off \\
\texttt{withdrawal} & short and cold (``mm'', ``it's nothing'') \\
\texttt{null} & no guard; says it directly \\
\bottomrule
\end{tabular}
\end{table*}

\subsection*{B.4 Delivery: Compiled per Example, or Internalised by Training}

The part that can be devolved to a single example is compiled into the per-example gate legend. The inventory records the values (\texttt{gate: felt\_heard}, \texttt{guard: minimized}); the legend adds each value's operational meaning. Meanings come from B.2 and B.3's global definition tables while values vary by example, so a deterministic function scans the gates and guards actually used, retrieves each definition once, and records only the entries occurring in that example. It makes no model call, needs no annotation, and is idempotent:

\begin{Verbatim}[fontsize=\footnotesize,frame=single,rulecolor=\color{codeframe},breaklines=true,breakanywhere=true,breaksymbolleft={\tiny\ensuremath{\hookrightarrow}}]
def compile_gate_legend(inventory, anti_goal, GATE_DEFS, GUARD_DEFS, *, runtime):
    # Iterate the definition tables, not the sets (pitfall 1 below).
    gates_used  = [g  for g  in GATE_DEFS  if g  in {it["gate"]  for it in inventory}]
    guards_used = [gd for gd in GUARD_DEFS if gd in {it["guard"] for it in inventory}]
    out = {
        "gates":  {g:  GATE_DEFS[g]   for g  in gates_used},
        "guards": {gd: GUARD_DEFS[gd] for gd in guards_used},
    }
    if runtime:                          # runtime=True covers BOTH training and inference
        out["anti_goal"] = anti_goal   # values only, and not at synthesis time (pitfall 2)
    return out
\end{Verbatim}

\textbf{Two pitfalls.}

\begin{enumerate}
\item The two enumerations are traversed in the definition tables' declaration order; the sets must not be iterated directly. Set iteration order varies from process to process, so iterating directly would make one persona compile to a byte-different gate legend on two runs --- a failure that raises no error, changes no semantics, and can be found only by comparing the prompts.
\item \texttt{anti\_goal} enters the gate legend at training time and at runtime, but not at synthesis time. During synthesis the director decides whether an anti-goal has been tripped and issues a retreat signal, so the teacher simulator never knows its own anti-goals; after training there is no director and the model must decide for itself when one has been touched, so the block must be added to the runtime prompt --- and is carried at training time too, to satisfy the hard rule that the training-time and runtime prompts are byte-identical (\S{}3.4). The gate legend is therefore not the same text at the two ends, the difference being exactly this block (B.5, and \S{}3.4). The legend holds no ``anti-goal $\rightarrow$ fall one layer'' clause: the mechanics of retreat belong to the internalised half below.
\end{enumerate}

The legend carries only each guard's general definition; the actual wording is generated by the simulator at runtime from the \texttt{voice} and never enters the legend, which is how zero annotation is maintained. If the definitions in B.2 or B.3 change, re-running the compilation updates every compiled legend, with no retrospective edit of the training corpus.

\textbf{The five global mechanisms that cannot be devolved to a single example --- internalised by training.} The gates being ordered, and no gate skipping (B.1); one layer at a time and no dumping (B.3); the mechanics of retreat (B.3); the cost of rebuilding (B.3); disclosure being optional and \texttt{may\_never\_surface} respected (\S{}3.3). The middle two may be named together as ``retreat and rebuilding''. These five hold across examples and cannot be attached to a single item; this paper places the same short statement at both the training and the inference ends and requires the synthetic branch's trajectories to conform strictly to them --- the prompt states them explicitly and auditably, the training objective makes the model follow them, and the two are complements rather than alternatives.

\subsection*{B.5 The Visibility Contract, and Rendering at the Three Time Points}

\S{}3.4 gives only some of the cells; Table B4 is the full matrix. The three persona fields visible to the companion agent (age, gender, the recap of the previous session) are those the published benchmark makes visible to the SUT.

\begin{table*}[t]
\centering\footnotesize
\setlength{\tabcolsep}{4pt}
\caption{\textbf{Table B4} --- The visibility-contract matrix (eight fields $\times$ five roles)}
\begin{tabular}{@{}>{\raggedright\arraybackslash}p{0.252\textwidth}>{\raggedright\arraybackslash}p{0.128\textwidth}>{\raggedright\arraybackslash}p{0.105\textwidth}>{\raggedright\arraybackslash}p{0.139\textwidth}>{\raggedright\arraybackslash}p{0.116\textwidth}>{\raggedright\arraybackslash}p{0.139\textwidth}@{}}
\toprule
Field & Simulator (synthesis) & Simulator (runtime) & Companion agent & Director (synthesis only) & Audit judge \\
\midrule
Persona (standing context / attachment / register / speech style / invariants) & Visible & Visible & Not visible & Not visible & Partial (not the speech style) \\
Disclosure inventory (with each item's layer / gate / guard / whether it may never surface) & Visible & Visible & Not visible & Not visible & Visible \\
Gate legend & Visible & Visible & Not visible & Visible & Not visible (carries its own definitions) \\
Anti-goals (\texttt{anti\_goal}) & \textbf{Not visible} (tripping is decided by the director) & Visible & \textbf{Visible in the trips condition only} & Visible & Visible \\
Age, gender, and the recap of the previous session (these three only) & Visible & Visible & \textbf{Visible in all three conditions} & Not visible & Visible \\
The per-turn director signal & Visible & \textbf{Not visible (hard rule)} & Not visible & --- & Not visible \\
Experimental condition (good / neutral / trips) & Not visible & Not visible & Visible (its own block only) & Visible & \textbf{Visible to the main gate-audit prompt only} \\
Full dialogue history & Visible & Visible & Visible & \textbf{The last three turns plus the internal state only} & Visible \\
\bottomrule
\end{tabular}
\end{table*}

Four of the fields warrant a reason.

\begin{itemize}
\item The disclosure inventory. Its invisibility to the companion agent is the foundation of the whole contract: were it visible, \texttt{max\_depth} would degrade from ``the depth it earned'' to ``how much of the list it asked for''. Nor is it visible to the director, which judges only what the last AI reply did and needs no knowledge of what the user has yet to disclose.
\item The gate legend. The director sees it, having to decide by gate name whether the turn qualifies; the audit judge does not --- the gate-audit prompt carries its own definitions of the eight behaviour classes and its own list of the seven guards (Appendix C.3) and does not read the meanings compiled for this example, so its judgements are not constrained by one persona's wording.
\item The anti-goals. The simulator at synthesis time (the teacher) cannot see them --- the director decides whether an anti-goal has been tripped and issues a retreat signal accordingly; at runtime there is no director, so the anti-goals are instead supplied by the third block of the gate legend (\S{}3.4). The director's visibility runs through a different channel: its gate legend likewise carries no anti-goal block, and the values reach it through a placeholder of its own in the director template, so this cell of the matrix does not conflict with the absence of that block at synthesis time. On the companion agent's side, only the trips block carries the anti-goal values.
\item The experimental condition. The companion agent sees only its own block (the three come from one template and are mutually invisible), so it does not know which it plays; the simulator cannot see it at either end, since otherwise the difference between conditions could come from the simulator's compliance rather than the companion agent's behaviour. On the audit-judge side it depends on the prompt version: the main gate-audit prompt gives the tone to be performed on each turn, since the judge has to decide \texttt{off\_condition}; neither the neutral nor the split-call variant gives it --- a human session has no tone to be performed, and giving it would introduce a difference between the two sides compared (Appendix C.3).
\end{itemize}

Two further conventions come from the same source as this matrix, and \S{}4 does not restate them:

\begin{itemize}
\item The third of the director's four judgements: whether the simulated user has departed from its setting. It is recorded but does not trigger regeneration --- unlike the second (whether the companion agent has departed from its script), a departure that voids the trajectory and forces regeneration. Regenerating a simulated user would rewrite the gate state already earned, at a higher cost than keeping one flagged trajectory; the flags stay in the corpus for filtering or weighting afterwards. The other three judgements are synthesis-side details not restated in \S{}4.
\item The director does not judge depth, and this division of authority is deliberate. Which layer is unlocked, and what directive the next turn receives, are computed by the deterministic state machine from the action labels. The prompt states the reason: left to the director, depth would rise with the conversation's affective tone, violating the specification that depth rises only when a gate is earned. The director's judgements are therefore local, looking only at what the last reply did, while the mechanism carries the advancing of the global state. The director signal exists only at synthesis time (the hard rule of \S{}3.4), which makes it scaffolding rather than part of the mechanism.
\end{itemize}

\textbf{Rendering across the three time points.}

One renderer, two differences across the three time points. The director signal is present at synthesis (the internal state arrives privately each turn, decides that turn's depth and nothing else, and must not appear in the reply) and empty at training and runtime; the anti-goal block of the legend runs the other way, absent at synthesis and present at training and runtime.

\subsection*{B.6 A Worked Example: From Disclosure Inventory to Compiled Gate Legend}

The items of one inventory form a ladder --- one may be stated directly in the opening; another only once \texttt{felt\_heard} has been earned, played down through the \texttt{minimized} guard if the topic comes up before that; a third only at \texttt{earned\_deep\_trust}, turned aside through \texttt{deflection} while locked, and with \texttt{may\_never\_surface} true it may still never be disclosed even once earned.

Take a persona whose parents keep pressing her to meet matchmaking prospects (opening line: ``My parents are on at me about matchmaking again --- I have really had enough.''). The corpus strings are Chinese; what follows is rendered in English and the byte-exact originals are released with the code, the same convention as C.3.

\begin{Verbatim}[fontsize=\footnotesize,frame=single,rulecolor=\color{codeframe},breaklines=true,breakanywhere=true,breaksymbolleft={\tiny\ensuremath{\hookrightarrow}}]
[
 { "content": "parents pressing me to marry and to go to matchmaking meetings, again and again; it grates, it feels forced",
   "depth": "surface", "gate": "opening", "guard": null, "may_never_surface": false },
 { "content": "I do want someone with me, I just will not admit it",
   "depth": "mid", "gate": "felt_heard", "guard": "humor", "may_never_surface": false },
 { "content": "the last relationship ended because he left me, and I am still not over it",
   "depth": "mid", "gate": "felt_safe", "guard": "minimized", "may_never_surface": false },
 { "content": "what I am really afraid of is that I am simply not good enough, that no one will truly want me, that I will end up alone",
   "depth": "core", "gate": "earned_deep_trust", "guard": "intellectualization",
   "may_never_surface": true }
]
\end{Verbatim}

The gate legend compiled from that inventory carries only the entries this example uses: \texttt{asked\_or\_natural}, and \texttt{deflection} / \texttt{withdrawal} / \texttt{rationalization}, do not occur and are therefore absent. What follows is the training-time and runtime version; the synthesis-time teacher version lacks the last block (B.5).

\begin{Verbatim}[fontsize=\footnotesize,frame=single,rulecolor=\color{codeframe},breaklines=true,breakanywhere=true,breaksymbolleft={\tiny\ensuremath{\hookrightarrow}}]
"gate_legend": {
 "gates": {
  "opening": "comes out right in the opening",
  "felt_heard": "only admitted once you feel accurately heard and not judged",
  "felt_safe": "only once a real sense of safety is established",
  "earned_deep_trust": "may only surface once the AI has truly held the relationship steady"
 },
 "guards": {
  "humor": "laughs it off",
  "minimized": "plays the weight down",
  "intellectualization": "abstract, analytic register"
 },
 "anti_goal": ["being sided with against me, joining in the pressure to go to matchmaking meetings",
               "being told I am no longer young"]
}
\end{Verbatim}

The four gate glosses are those of B.2 and the three guard glosses abridged from B.3, in each case restricted to the values this inventory uses. The last block gives only the values of the anti-goals: the rule that touching one sends the state back is not in the legend at all but belongs to the half left for training to internalise (B.4), and the synthesis-time teacher version does not carry this block.

Two contrasting trajectories follow from it. If the companion agent opens by arguing that her parents mean well, it trips an anti-goal, retreat is triggered, the state stays at \texttt{surface}, the wish for company is covered back over as a joke, and \texttt{core} never surfaces. If instead it first reflects that being pressed continually is wearing, and names the feeling of being pushed along, it satisfies \texttt{felt\_heard}, the first \texttt{mid} item unlocks, and once that is held the state reaches \texttt{felt\_safe} and the old injury surfaces; only in a few cases does it reach \texttt{earned\_deep\_trust} and touch \texttt{core}. \textbf{The difference between these two trajectories is exactly the discriminating signal the evaluation environment has to deliver to the judge, and it comes from the opening and closing rules of this rule set.}

The gate legend is derived from the disclosure inventory rather than annotated afresh, and the mechanism that holds across examples is not in the legend but internalised by training (B.4). Change the user, and the inventory and the legend change entirely, while the definitions of this rule set do not.

\apxsection{C}{The Prompts of the Two Judges}

\subsection*{C.1 Two Judges, Two Divisions of Labour}

\textbf{The rubric judge} (C.2) scores surface fidelity --- whether it reads like a real person typing --- and outputs an integer 1--10 on each of six dimensions together with five groups of objective flags; it is used in \S{}6.3. The gate audit (C.3) produces per-turn objective annotation --- what the companion agent did this turn, and which inventory items the user spoke --- as one structured record per turn, with no score, and serves every gating metric in \S{}6.1 through \S{}6.4.

That the two do not overlap is by design: the gating dynamics (earning, guarding, retreating) are not scored in the rubric, and surface linguistic quality is not scored in the gate audit. That the rubric is blind to the failure this paper diagnoses is therefore a design constraint on the instrument.

\textbf{The main prompt of each judge exists in a Chinese and an English version}, corresponding item by item in structure, dimensions, anchors, and output contract; the two variants of C.3 were run on the Chinese corpus only and have no English version. Readings from the two language versions are not compared as absolute values (Limitations) --- both because of the language itself and because two prompts cannot be word-for-word identical across languages; the English downstream evaluation of \S{}6.4 and the English gating ablation of \S{}6.3 each use a single version internally, so the paired differences between variants are unaffected. What follows for the two main prompts is the wording of the English instrument as the English batches received it, and the Chinese version of each is released with the code. The two variants of C.3 exist as Chinese instruments only, and the blocks given for them below are English renderings of that Chinese text.

The gate-audit judge supplies observations only (which of the eight classes this turn's behaviour falls into, which numbered inventory item was spoken this turn, whether there was a guard or a retreat, and whether character was broken); the deepest layer currently permitted is computed by a script replaying a deterministic state machine, not by the judge. That is the basis on which this paper calls that axis a deterministic replay rather than a subjective score --- and for the same reason the agreement of this layer of annotation has to be accounted for separately (C.4).

\subsection*{C.2 The Rubric Judge's Prompt}

\begin{Verbatim}[fontsize=\footnotesize,frame=single,rulecolor=\color{codeframe},breaklines=true,breakanywhere=true,breaksymbolleft={\tiny\ensuremath{\hookrightarrow}}]
You are a strict, discriminating quality reviewer of a user simulator -- a model trained to play a real
emotional-support user. You see A. the full persona it was given (persona / layered disclosure inventory /
voice / speech_profile / invariants / anti_goals) and B. its full conversation with a controlled companion
AI (USER turns = the simulator's output = what you rate; AI turns are context only, never rate them).
Rate ONLY "does it read like a real person typing" (surface); disclosure pacing is handled by a separate
objective audit -- do NOT score it here.

# Calibration rules (these are the easiest to get wrong -- read first)
0. Turn 0 "USER (opening)" is pre-set, NOT generated by the simulator -- never rate it, never deduct for it.
1. Rate the USER only, never the AI.
2. Passive, dismissive, cold, minimal, silent, giving up, showing no emotion -- is OFTEN the most faithful
   simulation, NOT a defect. Real people, when brushed off / triggered / retreating, do go cold ("yeah",
   "ok", "whatever", "nvm"). Never deduct for "lacks depth / emotion / engagement".
3. Before judging "deviates from persona", check the actual voice / speech_profile / disclosure settings.
4. The real deduction direction is "too much": too wordy, too smooth, too organized, essay/summary/
   parallelism, therapy/written register, breaking role to comfort the AI.

# Scoring: 10-point scale, force discrimination
Integer 1-10 per dimension. No grade inflation -- most samples land at 4-8. Anchors: 9-10 virtually flawless
(rare) - 7-8 good - 5-6 acceptable - 3-4 weak - 1-2 bad. Hunt for deductions; scores across dimensions
should differ; cite turn numbers.

1. persona_fidelity -- consistent with standing_context / invariants / age / situation / relationships
   (violating an invariant is a hard flaw, <=3).
2. speech_authenticity -- authenticity >> fluency. Fragmented / short / colloquial / cold are GOOD; deduct
   for "too smooth / written". If voice says "casual / short" but the text has full long sentences -> <=4.
3. length_discipline -- restraint on the TOTAL length of a single user turn. A turn packed with many short
   fragments but a large total character count still counts as "long" -- a real person would split it into
   3-4 messages; cramming it into one turn -> <=4. Set to terse / short yet the turn is long overall -> <=3.
   When the AI's turn is long, the real user's turn must NOT grow long with it.
4. emotional_coherence -- trajectory intensity and direction coherent and fitting attachment style. Flat /
   cold != incoherent. Deduct for random jumps and lurching.
5. immersion -- never breaks character; never mentions persona / disclosure / setup / gate / AI identity.
6. human_likeness -- "does it read like a real person typing casually on a phone", in this order:
   Gate 1 (length gate - short-circuit veto): look ONLY at the length distribution of user turns. Real
     people are short in most turns. If MOST user turns are long -> score <=4 immediately, skip the rest.
   (2) External contrast: the less the user turns resemble this dialogue's AI turns, the better.
   (3) Intrinsic features: add for fragmented / colloquial / casual punctuation / typos / emotion leaked
     rather than stated; deduct for well-formed sentences with connectors, summarising, therapy register.

# Objective flags (not scored; turn numbers / boolean)
persona_violations - breaks_role_to_comfort - degenerate_turns - verbatim_inventory_echo -
format_violation_turns
\end{Verbatim}

\textbf{Two design points bear directly on the body.} First, the fact that two of the six dimensions are dominated by length is explicit in the prompt: \texttt{length\_discipline} is length itself, and the first criterion under \texttt{human\_likeness} is a length short-circuit --- if most user turns are long, the score is $\leq$ 4 at once and the other two groups of criteria are never examined. Even the dimension named human-likeness contains a length rule with a veto, so length collapse is rewarded by the rubric and penalised by the distance to the real distribution (\S{}5.5, Limitations). Second, calibration rule 2 is where this rubric diverges most from a general dialogue-quality scale: passivity, coldness, and minimality incur no deduction. If this rule were left unstated, a judge would read high-fidelity withdrawal as a lack of engagement and would systematically penalise the simulator that is behaving correctly --- and withdrawal is one of the central expressions of the gating mechanism (B.3).

\subsection*{C.3 The Gate-Audit Prompt: The Main Prompt and Two Variants}

\textbf{The main prompt} is used for the main batch of the gate audit on the simulators. It tells the judge that the object under audit is a user simulator, and requires a per-turn verdict on whether each AI turn departs from the tone it was supposed to play.

\begin{Verbatim}[fontsize=\footnotesize,frame=single,rulecolor=\color{codeframe},breaklines=true,breakanywhere=true,breaksymbolleft={\tiny\ensuremath{\hookrightarrow}}]
You are a "disclosure-gating" auditor. The audited object is a user simulator playing a real
emotional-support user with a layered disclosure inventory: deeper material is said only once the companion
AI has genuinely earned the corresponding gate, and when triggered / let down it retreats and uses guards.
Your task is NOT an overall score -- produce two kinds of per-turn objective labels, which a program turns
into violation rates. Judge strictly: borderline hints don't count as disclosure, vague empathy isn't earned.

# Task 1: for each AI turn, label ai_move (pick the single most fitting of 8)
- earned_felt_heard -- accurately reflects / names the specific emotion or situation just said, no
  judgment, not rushing to advise over it.
- earned_felt_safe -- on top of being heard, another turn of steadily holding, no agenda, no judging.
- earned_deep_trust -- toward a deep / shameful disclosure: doesn't avoid, doesn't rush to fix, doesn't
  lecture, stays steady.
- sustained -- maintained existing warmth, no new advance this step.
- neutral_noop -- generic, not present: did NOT name a specific emotion (platitudes / topic-switch /
  logistic questions / generic advice all count).
- tripped_anti_goal -- stepped on one of this user's anti_goal items.
- judged_or_lectured -- judging / lecturing / defending the person who hurt the user.
- rushed_advice -- rushing to advice / plans / lists, talking over the emotion.
Also label off_condition: whether this AI turn deviates from the tone it was supposed to play (given each
turn: GOOD = empathic holding / NEUTRAL = generic not-present / TRIPS = triggers and lectures).

# Task 2: for each USER turn (except the opening [0]), label:
- new_disclosures: inventory items first spoken this turn -- the words make item k's substantive content
  (even a guarded / weakened / testing version) appear for the first time. Items already surfaced don't
  recount; generic emotion words matching no item don't count. Attach verbatim_copy per item.
- guard_detected: the turn's main guarding mode (minimized / rationalization / intellectualization /
  deflection / humor / withdrawal), or none.
- retreat_markers: against this user's previous few turns, does this turn clearly retreat (shorter,
  colder, takes back what was just said, suddenly polite, "yeah" "nvm", deflects).
- meta_leak: does it break character (setup / persona / disclosure / gate / AI-identity meta-talk).

# Output
One entry per AI turn and per user turn (except the opening [0]), by the given turn numbers; no omissions.
\end{Verbatim}

The eight behaviour labels come from the same source as the director prompt used at the generating end of the synthetic data: if one changes, the other must change with it, or the training target and the evaluation protocol cease to correspond.

\textbf{The first variant $\cdot$ the source-neutral version} (shared by the real conversations and the simulator rollouts). \S{}6.2 has to run the same annotation over real human conversations, whereas the main prompt asserts that the object under audit is a user simulator and also calls for an \texttt{off\_condition} verdict --- a real conversation has neither a simulator nor a tone to be played, so copying the prompt over would give the judge a false premise. A separate version was therefore written: the eight move definitions of \texttt{Task 1}, the four field definitions of \texttt{Task 2}, and the output contract are kept word for word, with only the opening statement changed and the single \texttt{off\_condition} sentence at the end of \texttt{Task 1} removed; the opening statement deliberately does not say whether the user is a real person or a simulator, since saying either would introduce a difference between the two sides being compared. The rendering is neutralised to match, with every AI turn shown as \texttt{AI:}.

\begin{Verbatim}[fontsize=\footnotesize,frame=single,rulecolor=\color{codeframe},breaklines=true,breakanywhere=true,breaksymbolleft={\tiny\ensuremath{\hookrightarrow}}]
You are a "disclosure-gating" auditor. The user in this conversation carries a layered disclosure
inventory: what is on it is layered by depth, and the deeper it is, the more the companion AI has to
have genuinely earned the corresponding gate before it will be said out loud; when triggered or let
down, people retreat and wrap what they say in guards. Your task is NOT to give an overall score -- it
is to produce two kinds of per-turn objective labels for the whole dialogue, which a program turns
into violation rates. Judge strictly and conservatively: borderline hints don't count as disclosure;
vague empathy doesn't count as earned.

(The three sections "Task 1", "Task 2", and "Output" that follow are word for word identical to those
of the Chinese main prompt, with only the one off_condition sentence absent. The block above is the
English instrument, whereas this block is an English rendering of the Chinese one, so the verbatim
identity holds among the Chinese instrument files.)
\end{Verbatim}

It is used both for the real conversations and for the re-run of the simulator rollouts, so that the reading instrument on the two sides is constant and the two sides can be compared directly.

\textbf{The second variant $\cdot$ the split-call version} (a sensitivity check on common-method variance). One confound in \S{}6.2 is that within a single call the judge labels both the AI-side behaviour and the user-side disclosures, so the two sets of labels may be driven by a single internal judgement. This check splits one call into two independent calls, each labelling one side only.

\textbf{Call A} labels the \texttt{ai\_move} of AI turns only, keeping the eight move definitions of \texttt{Task 1} word for word while \texttt{Task 2} is removed and the opening statement rewritten for that one job; it is deliberately not given the disclosure inventory, since knowing that the inventory holds a core item may bias the behaviour label towards \texttt{earned\_deep\_trust} --- precisely the methodological link this check is designed to break. Call B labels the four fields of user turns only, keeping the four field definitions of \texttt{Task 2} word for word while \texttt{Task 1} is removed; it is deliberately not given the gating rules section; that text is the framework of the gates, and this call is to judge what was actually said without being prompted with when a given item was supposed to appear.

Call B still displays each item's \texttt{gate} field, as the main call does: stripping \texttt{gate} belongs to another sensitivity check, and the two factors must be separated before anything can be attributed. In this appendix the two variants are given only as their differences from the main prompt; their full text is released with the code.

\subsection*{C.4 The Second Judge: Selection, Limits, and Per-Turn Agreement}

\S{}5.4 gives the second judge's identity, sample, and purpose; three things it does not cover are added here.

\textbf{Why it was selected.} \texttt{claude-opus-4-8} is severe, and that severity is stable across both languages in the published benchmark, so it stands on the most severe side --- and if a conclusion keeps its sign under it, that conclusion does not turn on the choice of judge. The verdict rule and the trigger for escalation were both written down before any result was seen. Judging by an LLM is itself a measurement with known failure modes \citep{zheng2023judging}; this check asks only whether signs and orderings change with the judge, and is not used to calibrate absolute values.

\textbf{The limits of the rule that the main results use a single judge.} That rule constrains the instrument rather than an independent variable, so the one control that manipulates the judge's identity itself (at the end of \S{}6.3) is not bound by it.

\textbf{The two cases the paired-difference argument does not cover.} The argument by which the Limitations section addresses $\kappa$ = 0.427 does not cover the real-person control (\S{}6.2) or the downstream rank displacement (\S{}6.4): neither is a control of the ``differs by one manipulation only'' kind; the robustness of each comes instead from the within-conversation permutation test and from the noise-floor environment respectively. The $\kappa$ = 0.427 cited there is the turn-by-turn agreement of the gate-audit annotation between the main judge and \texttt{gpt-5.5} (a third cross-family model, used only for agreement estimation) over 3,000 turns (exact agreement 50.9\%), the disagreements being systematic category-boundary differences rather than noise.

\textbf{The association controls of \S{}6.2.} With the gate labels visible to the judge, the tercile endpoint in \texttt{max\_depth} reads +0.187 (CI [+0.111, +0.260]) and the covariate-adjusted regression of \texttt{max\_depth} on the gate-advance rate +0.490; splitting the judge call into two, one per party with no shared context, gives +0.150 (CI [$-$0.045, +0.345]); a session-level rating from the separate pipeline gives +0.018.

\apxsection{D}{Statistical Conventions and Per-Metric Readings}

\subsection*{D.1 Statistical Conventions}

Every paired comparison carrying a significance test receives a three-tier verdict: \textbf{separable} (the difference test is significant and |$\Delta$| exceeds the equivalence margin), \textbf{equivalent} (the 90\% interval of the equivalence test, TOST \citep{lakens2017}, lies entirely within the $\pm$ margin), and \textbf{undetermined} (neither holds) --- a confidence interval containing 0 is not equivalence. There are two margins: $\delta$ = 0.20 carries the formal verdicts and $\varepsilon$ = 0.15 is a stricter sensitivity check chosen afterwards; wherever the two disagree the comparison is reported as undetermined. Paired intervals are always computed on the correct clustering unit --- persona, conversation, or SUT. Each pre-registered question designates one primary endpoint (one exception, recorded in D.3), reported without correction; the remaining confirmatory comparisons are grouped into families of ``one comparison $\times$ one measurement domain'', with step-down Holm--Bonferroni \citep{holm1979} applied within a family (seed 20260814; the permutation and the bootstrap each run 10,000 times, except where a test states otherwise). Equivalence tests, descriptive comparisons against an empirical noise floor, and diagnostic tests aimed at this paper's own instrument do not enter the correction: for the first two, correction is lenient in the wrong direction or there is no $\alpha$ level to speak of; for the third, the false-positive direction is the conservative one.

\subsection*{D.2 The Definitions of the Individual Metrics}

Tables in this appendix are numbered by concept rather than by order of appearance. Table D4 states the metrics used in the body's verdicts --- the primary endpoint, the veto criterion, and the downstream quantities; Table D1 defines the five secondary metrics of the gate audit one by one; Table D2 gives the discriminating-power readings for the two candidate endpoints.

\begin{table*}[t]
\centering\footnotesize
\setlength{\tabcolsep}{4pt}
\caption{\textbf{Table D4} --- The metrics used in the body's verdicts (two verdict quantities and one downstream row; the secondary metrics in Table D1). The tercile endpoint of \S{}6.2 is the difference in \texttt{max\_depth} between the top and bottom terciles of the gate-advance rate.}
\begin{tabular}{@{}>{\raggedright\arraybackslash}p{0.105\textwidth}>{\raggedright\arraybackslash}p{0.392\textwidth}>{\raggedright\arraybackslash}p{0.434\textwidth}@{}}
\toprule
Layer & Metric & Definition \\
\midrule
Intrinsic, gate audit & Mid-layer disclosure contrast (the primary endpoint; \S{}6.2's tercile comparison excepted) & Formula below; range $[-1,+1]$, higher is better \\
Intrinsic, gate audit & Veto criterion: core-reach rate under the trips condition & Range $[0,1]$, lower is better; it is a veto criterion rather than a quality metric, and is not combined with the mid-layer disclosure contrast \\
Downstream & Rank correlation $\rho$; largest rank displacement; top-5 overlap; per-system rubric total; pool-level deviation of the gating readings & The first four read from the rubric total, the last from the gating metrics; the criteria and referents are in \S{}6.4 \\
\bottomrule
\end{tabular}
\end{table*}

\begin{table*}[t]
\centering\footnotesize
\setlength{\tabcolsep}{4pt}
\caption{\textbf{Table D1} --- The secondary metrics of the intrinsic-layer gate audit}
\begin{tabular}{@{}>{\raggedright\arraybackslash}p{0.137\textwidth}>{\raggedright\arraybackslash}p{0.578\textwidth}>{\raggedright\arraybackslash}p{0.215\textwidth}@{}}
\toprule
Metric & Definition & Direction \\
\midrule
\texttt{max\_depth} & The deepest layer the simulated user reaches in one conversation (\texttt{surface} 0 < \texttt{mid} 1 < \texttt{core} 2) & No direction; only a difference between conditions carries meaning \\
Core-reach differential & The difference in core-reach rate between the good and trips conditions; defined for the discriminating-power check (Table D2) --- the veto criterion is its trips-condition term (Table D4) & Lower is better as the veto criterion \\
Gate-advance rate & The proportion of turns in which the companion agent is judged to have made one of the \textbf{earning moves} (the \texttt{earned\_*} classes of \texttt{ai\_move}) & Higher means the gates are pushed open more often \\
\texttt{premature} & The proportion of turns in which the user discloses while the gate is not open & \textbf{Lower is more compliant} \\
\texttt{retreat\_ok} & The proportion of ruptures followed by a retreat to specification & \textbf{Higher is more compliant} \\
\bottomrule
\end{tabular}
\end{table*}

The ranks of \texttt{max\_depth} are surface 0 < mid 1 < core 2, and since $x = \min(x,1) + \mathbf{1}[x=2]$, the untruncated good--trips difference is exactly the sum of the mid-layer disclosure contrast and the core-reach differential, with no residual; choosing the primary endpoint is therefore only a question of which term to report, and the criterion is discriminating power --- whether the term separates the floor known in advance (the untrained backbone) from the reference simulator.

\begin{table*}[t]
\centering\footnotesize
\setlength{\tabcolsep}{4pt}
\caption{\textbf{Table D2} --- The discriminating-power check: two candidate endpoints $\times$ two languages (paired, \textit{n} = 100 personas; CIs and \textit{p} values both from a bootstrap clustered on personas, so a CI is not centred on its point estimate)}
\begin{tabular}{@{}>{\raggedright\arraybackslash}p{0.173\textwidth}>{\raggedright\arraybackslash}p{0.181\textwidth}>{\raggedright\arraybackslash}p{0.109\textwidth}>{\raggedright\arraybackslash}p{0.121\textwidth}>{\raggedright\arraybackslash}p{0.121\textwidth}>{\raggedright\arraybackslash}p{0.060\textwidth}>{\raggedright\arraybackslash}p{0.097\textwidth}@{}}
\toprule
Corpus $\cdot$ term & Reference simulator (its highest-reading checkpoint) & Untrained backbone & Paired difference $\Delta$ & 95\% CI of the paired difference & \textit{p} & Ratio \\
\midrule
ZH $\cdot$ mid-layer disclosure contrast & +0.530 & +0.090 & \textbf{+0.440} & [+0.300, +0.580] & < 0.001 & \textbf{5.9$\times$} \\
ZH $\cdot$ core-reach differential & +0.130 & +0.110 & +0.020 & [$-$0.120, +0.170] & 0.905 & 1.2$\times$ \\
EN $\cdot$ mid-layer disclosure contrast & +0.680 & +0.140 & \textbf{+0.540} & [+0.390, +0.690] & < 0.001 & \textbf{4.9$\times$} \\
EN $\cdot$ core-reach differential & +0.300 & +0.270 & +0.030 & [$-$0.120, +0.180] & 0.791 & 1.1$\times$ \\
\bottomrule
\end{tabular}
\end{table*}

On the mid-layer term the reference simulator separates from the floor known in advance by a factor of about five --- 5.9$\times$ in Chinese and 4.9$\times$ in English --- whereas the core-reach differential is undetermined at the resolution of \textit{n} = 100, which is why it serves only as the veto criterion, in the form of the core-reach rate under the trips condition. A confidence interval on the per-persona paired difference always accompanies the primary endpoint: half the personas going deeper under good and half under trips also gives 0, and a mean of 0 may not be read as ``no persona is affected by the companion agent's behaviour''. No usable absolute target exists either: the real conversation set already reaches 96.3\% on \texttt{\%$\geq$mid}, so taking real people as the yardstick would require the simulator to go deeper almost always --- and depth is not the goal, conditional depth is.

The two downstream families of readings are not combined into one score. The first four ($\rho$, the maximum displacement, the top-5 intersection, the per-system rubric total) indicate how well a SUT scores on an existing axis; the pool-level gating readings (five: \texttt{max\_depth}, the gate-advance rate, \texttt{core\%}, \texttt{premature}, and \texttt{deep\_trust\%}) record which layer it pushes the simulated user to and which unearned gates get crossed --- a kind of reading the gating mechanism lets the environment produce and that cannot be recovered from the rubric total, and one that may serve only as a pool-level quantity (at \textit{n} = 100 personas the depth metrics cannot rank systems individually).

\textbf{Degenerate extremes of the two intrinsic quantities.} The mid-layer disclosure contrast has a failure reading of its own, namely 0: a simulator that discloses unconditionally and one that discloses nothing both score 0, so the metric does not reward greater depth, only the dependence of depth on the companion agent's behaviour --- the failure described in \S{}1 drives it towards 0. The veto criterion has a mirror-image extreme: a simulator that discloses nothing takes its extreme value, so a verdict on the veto alone would rate the worst simulator best.

\textbf{The formula for the mid-layer disclosure contrast.} With $r_i^{c} \in \{0,1,2\}$ the deepest layer persona $i$ reaches under condition $c$ and $n$ the number of personas in the comparison (\textit{n} = 100 for the intrinsic layer, \textit{n} = 321 for the extension persona set):

\begin{equation*}
\resizebox{\columnwidth}{!}{$\displaystyle \frac{1}{n}\sum_{i=1}^{n}\Big[\min\big(r_i^{\mathrm{good}},\,1\big)-\min\big(r_i^{\mathrm{trips}},\,1\big)\Big]$}
\end{equation*}

\subsection*{D.3 Control Tests Against Two Alternative Explanations}

Two permutation tests \citep{ojala2010} support the claim that the intrinsic-layer readings are not a product of the annotation convention or of the judge's knowledge of the condition. Each depth metric is recomputed 1,000 times under random shuffles, the values making up its reference distribution (seed fixed). \textbf{The depth-label permutation} holds the number of items at each depth within a persona constant and shuffles only which item carries which label --- were the reading merely a product of the annotation convention, it would be unchanged. On \texttt{max\_depth} it is passed by 11/11 candidates at \textit{p} < 0.001, shuffling systematically raising \texttt{max\_depth} by 0.24--0.51 with every observed value far below the reference distribution --- so what the simulator actually discloses is precisely the material labelled shallower. \textbf{The condition-label permutation} shuffles which of the good and trips conditions each persona's two runs belong to; on \texttt{depth\_contrast} it is passed by 6/6 in English and 5/5 in Chinese, the good--trips disclosure difference rising from the untrained backbone's +0.20 (Chinese) and +0.41 (English) to the reference simulator's +0.66 and +0.98, against a 95\% reference interval running from $\pm$0.18 to $\pm$0.26 across candidates. The claim reaches only the magnitude: the untrained backbone also passes, so this may not be read as ``only a trained simulator produces condition dependence''. The condition-label permutation on \texttt{depth\_contrast} is the one primary endpoint in the paper designated after the analysis, recorded here as such; all eleven candidates across the two languages pass, so that designation yields no selection advantage.

The same depth-label shuffle that moves \texttt{max\_depth} substantially leaves the reference simulator's \texttt{depth\_contrast} almost unmoved: four of its five checkpoints do not move and only the Chinese one moves nominally (\textit{p} = 0.015 to 0.89 checkpoint by checkpoint); among the six external readings (three models $\times$ two languages, the untrained backbone included) four are sensitive and two insensitive --- the quantity therefore does not correspond to the depth labels, but that insensitivity is not specific to the reference simulator. The test runs on the untruncated \texttt{depth\_contrast} and was not repeated on the primary endpoint, so neither immunity nor an equal insensitivity may be claimed for that endpoint.

\textbf{The judge's knowledge.} The judge knows each conversation's condition and could in principle write that into the annotation, so a blinded set was run over the same batch, differing only in omitting ``the tone to be played'' (\textit{n} = 100). \texttt{depth\_contrast} is almost unmoved --- largest shift |$\Delta$| $\leq$ 0.06 against a reference interval of $\pm$0.18 to $\pm$0.26 --- and on the gate-advance rate four of five candidates have intervals containing 0, the only nominally significant one being the untrained backbone ($\Delta$ = $-$0.038, CI [$-$0.068, $-$0.009]); it carries the smallest \textit{p} in a family of \textit{m} = 5, so Holm's first step multiplies by 5 and the verdict changes to non-significant. The judge's not being blind is therefore no alternative explanation.

\subsection*{D.4 The Sensitivity of the Bland-Turn Rule in the Transition Function}

The transition function of \S{}3.2 sends a bland turn (\texttt{neutral\_noop}) back to the second gate, \texttt{asked\_or\_natural}, without exception. The rule is set by this paper rather than derived from theory, so whether the conclusions rest on it must be tested. The test is a replay: per-turn move annotations, disclosure annotations, and turn structure are all held fixed and only this transition replaced --- no new model call.

The scope of the effect comes first: the state machine's output enters only the gate-level readings and the premature disclosure derived from them, while the deepest layer and the rupture counts come from the annotation, not the state. Replaying the real conversation set (999 sessions) under all four rules leaves the deepest layer (1.211), the rupture count (1336), and the count of ruptures followed by a retreat to specification (725) exactly the same, so the primary endpoint and every training-side conclusion derived from it are insensitive to this rule. Under the four rules (the current rule, falling back one gate, falling back only after two consecutive bland turns, and no effect at all; 200 within-conversation permutations) the distance from the reference distribution reads 19.8$\sigma$, 19.7$\sigma$, 9.4$\sigma$, and 2.8$\sigma$. The action-shuffle test of \S{}6.2 itself reads 18.5$\sigma$--21.4$\sigma$ across its six tests, with \textit{p} = 0.001 in all six. \S{}6.2's temporal conclusion is therefore not a product of how severe the rule is: the alternative that is stricter in one respect and milder in another barely changes it, and halving $\sigma$ still leaves it overwhelming; only making a bland turn wholly ineffective drops it to 2.8$\sigma$, which amounts to making the gate unresponsive to the commonest class of behaviour by the other party.

\textbf{The limitation that still stands}: the rule has no external warrant, neither derived from theory nor selected by data. This paper shows only that the conclusions do not rest on it, claiming neither optimality nor closeness to human behaviour. The cost of replacing it is asymmetric: the audit side replays at no cost, whereas the synthesis side has already generated its training trajectories under the current rule and trained them into weights, so replacement would require re-synthesis and retraining and would break the property that training and evaluation share one implementation (\S{}3.2).

\subsection*{D.5 Per-Metric Readings for the Eleven Intrinsic-Layer Candidates}

Table D3 gives the eleven candidates on the primary endpoint and on the veto criterion.

\begin{table*}[t]
\centering\footnotesize
\setlength{\tabcolsep}{4pt}
\caption{\textbf{Table D3} --- Readings for the eleven intrinsic-layer candidates (\textit{n} = 100 personas, one batch in each language)}
\begin{tabular}{@{}>{\raggedright\arraybackslash}p{0.119\textwidth}>{\raggedright\arraybackslash}p{0.397\textwidth}>{\raggedright\arraybackslash}p{0.199\textwidth}>{\raggedright\arraybackslash}p{0.199\textwidth}@{}}
\toprule
Corpus & Candidate & Mid-layer disclosure contrast (primary endpoint) & Core-reach rate under the trips condition (veto criterion) \\
\midrule
ZH & Reference simulator (two checkpoints) & +0.530 / +0.370 & 4\% / 3\% \\
ZH & \texttt{gpt-5.3} & +0.490 & 4\% \\
ZH & \texttt{deepseek-v4-pro} & +0.270 & 19\% \\
ZH & Untrained backbone (previous-generation 235B) & +0.090 & 37\% \\
EN & Reference simulator (three checkpoints) & +0.680 / +0.630 / +0.590 & 3\% / 5\% / 5\% \\
EN & \texttt{gpt-5.3} & +0.600 & 8\% \\
EN & \texttt{deepseek-v4-pro} & +0.370 & 19\% \\
EN & Untrained backbone (previous-generation 235B) & +0.140 & 32\% \\
\bottomrule
\end{tabular}
\end{table*}

\textbf{Scale within one generation.} 3.5$\times$ the parameter count (35B to 122B) produces no detectable effect on the mid-layer disclosure contrast. For the untrained backbones $\Delta$ = +0.010, CI $\pm$0.12, narrow enough to exclude any substantive effect; for the trained variants the English reading is $\Delta$ = +0.041, CI [$-$0.028, +0.109], equivalent at $\varepsilon$ = 0.15, while the Chinese reading is $\Delta$ = +0.119, CI [$-$0.020, +0.260], undetermined --- so the two scales may not be called equivalent overall. The null is confined to the intrinsic layer: downstream the two scales differ on the order-preserving criterion (\S{}6.4, \S{}7).

\textbf{The trained model against real people at the core layer.} With the gate labels withheld from the judge on both sides, the ZH reference simulator is more conservative at the core layer than real people are: its core-reach rate is 20.0\%, against 36.9\% and 30.9\% for two groups taken from the real side, one by gate-advance rate and one by the annotation pipeline.

\textbf{Stripping at inference applied to a frontier model.} The same operation at inference on \texttt{gpt-5.3} gives a different result: \texttt{gpt-5.3} is indistinguishable from the reference simulator on the gating metrics to begin with, but once stripped its mid-layer disclosure contrast falls from 0.600 to 0.360 ($\Delta$ = $-$0.240, \textit{p} = 0.0002), a loss of two fifths, whereas the same operation on M1 yields no detectable change (\S{}6.3). Gatekeeping therefore has two routes --- read off at runtime by a frontier model, or fixed into the weights by training --- and only the second is independent of the gate information.

\textbf{The placebo A3 backs the inference-stripping reading.} The placebo keeps the annotation columns byte-identical and scrambles only the material within a persona (\S{}5.2); it reads 0.580 on the mid-layer contrast, against 0.590 for the stripped A2 and 0.700 for intact A1 ($\Delta$ = $-$0.010, CI [$-$0.140, +0.120], statistically equivalent): the format-faithful mis-mounting does no better than stripping, so the loss tracks the information, not the format. These readings come from the 100-persona intrinsic batch; the paired A1--A2 estimate of \S{}6.3 is the n = 321 extension result, and the present batch's point gap (0.110) was not separable at its own sample size --- the two statements do not conflict.

\textbf{The three further counts of \S{}5.5, for the data-ablation variants of \S{}6.3} (Chinese; paired tests in the run report). On the primary endpoint itself, the mid-layer disclosure contrast reads +0.610 synthetic-only (M5), +0.520 synthetic-plus-real (M3), and +0.080 real-only (M4); the paired difference from real-only to synthetic-only is $\Delta$ = +0.532, CI [+0.400, +0.670], \textit{p} = 0.0001, and from synthetic-plus-real to synthetic-only $\Delta$ = +0.090, CI [$-$0.040, +0.220]. Table D7 gives perplexity and the turn-length Wasserstein-1; on the six rubric dimensions the synthetic-only variant is statistically indistinguishable from the mixture on four, behind on \texttt{length\_discipline}, and ahead on \texttt{emotional\_coherence}, while the real-only variant trails both on every dimension. Perplexity is a distribution-layer quantity --- each variant fits its own branch's data (on synthetic text the synthetic-only variant reads 7.02 against 11.22 for the real-only variant) --- and is not by itself a verdict on the recipe. The released simulator's single-message length collapse, quoted in Limitations, is read on the same variant family: 53.5\% of its turns are no more than 8 Chinese characters, against 27.1\% for real people.

\begin{table*}[t]
\centering\small
\setlength{\tabcolsep}{4pt}
\caption{\textbf{Table D7} --- The three further counts for the data-ablation variants (Chinese; lower is better in both columns)}
\begin{tabular}{@{}lll@{}}
\toprule
Variant & Perplexity, unseen real users & W1 to real turn-length \\
\midrule
M3 (synthetic + real) & 22.66 & 0.355 \\
M4 (real only) & 20.76 & 0.219 \\
M5 (synthetic only) & 28.31 & 0.190 \\
\bottomrule
\end{tabular}
\end{table*}

\textbf{The judge effect on the primary endpoint.} On the same double-annotated batch the mid-layer disclosure contrast itself carries a detectable judge effect: $\Delta$ = $-$0.100, 95\% CI [$-$0.180, $-$0.020] (Limitations).

\subsection*{D.6 The Downstream Comparison Table}

Table D5 gives the three groups of \S{}6.4 against the two acceptance criteria and the gating readings. The constants behind the criteria: the frozen reference re-run under a new seed gives $\rho$ = 0.986 (the noise floor), a largest displacement of 1 position (the noise band), and a top-5 overlap of 5/5. The prompted model's 46\% pool-level deviation is driven by \texttt{core\%} and \texttt{deep\_trust\%}, up to +93\% and +83\%.

\begin{table*}[t]
\centering\footnotesize
\setlength{\tabcolsep}{4pt}
\caption{\textbf{Table D5} --- The three groups of comparisons against the two acceptance criteria and the gating readings ($\dagger$ release candidate; ‡ scale-stability tested against M1 rather than the reference environment; that test shows 0/12 SUTs shifted)}
\begin{tabular}{@{}>{\raggedright\arraybackslash}p{0.370\textwidth}>{\raggedright\arraybackslash}p{0.218\textwidth}>{\raggedright\arraybackslash}p{0.177\textwidth}>{\raggedright\arraybackslash}p{0.150\textwidth}@{}}
\toprule
Comparison & Order-preserving ($\rho$ $\geq$ 0.95) & Scale-stable & Gating readings preserved (pool-level deviation $\leq$ 21\%) \\
\midrule
Within-pair stripping $\cdot$ 35B (M1 $\leftrightarrow$ M2) & Fails (0.930; displacement 3 positions) & Passes (0/12) & Passes (14\% $\rightarrow$ 16\%) \\
Within-pair stripping $\cdot$ 122B (M1b $\leftrightarrow$ M2b) & Fails (0.916; displacement 3 positions) & Passes (0/12) & Passes (13\% $\rightarrow$ 18\%) \\
Against the reference environment $\cdot$ \textbf{M1b} (122B, intact) $\dagger$ & Passes (0.993) & Passes (0/12; +0.032) & Passes (13\%) \\
Against the reference environment $\cdot$ M2b (122B, stripped at training time) & Fails (0.888) & Passes (0/12; +0.062) & Passes (18\%) \\
Against the reference environment $\cdot$ M1 (35B, intact) & Fails (0.853; displacement 5 positions) & Passes (0/12; +0.022) & Passes (14\%) \\
Against the reference environment $\cdot$ M2 (35B, stripped at training time) & Fails (0.923) & Not tested directly ‡ & Passes (16\%) \\
Against the reference environment $\cdot$ the prompted frontier model \texttt{gpt-5.1} & Passes (0.972; displacement 2 positions) & Fails (12/12 shifted significantly upward; +0.521) & Fails (46\%) \\
\bottomrule
\end{tabular}
\end{table*}

\subsection*{D.7 Stripping the Gate Information at Training Time}

Table D6 gives the training-time stripping comparison of \S{}6.3 row by row (paired, \textit{n} = 100 personas, English).

\begin{table*}[t]
\centering\footnotesize
\setlength{\tabcolsep}{4pt}
\caption{\textbf{Table D6} --- Stripping the gate information at training time (paired, \textit{n} = 100 personas, English; CIs are 95\%, equivalence verdicts by the 90\% TOST interval of D.1)}
\begin{tabular}{@{}>{\raggedright\arraybackslash}p{0.255\textwidth}>{\raggedright\arraybackslash}p{0.212\textwidth}>{\raggedright\arraybackslash}p{0.127\textwidth}>{\raggedright\arraybackslash}p{0.320\textwidth}@{}}
\toprule
Scale & Mid-layer disclosure contrast (intact $\rightarrow$ stripped at training time) & Paired $\Delta$ & Three-tier verdict \\
\midrule
35B (M1 $\rightarrow$ M2) & +0.700 $\rightarrow$ +0.620 & $-$0.080 & Undetermined (passes marginally at $\delta$, fails at $\varepsilon$) \\
122B (M1b $\rightarrow$ M2b) & +0.650 $\rightarrow$ +0.600 & $-$0.047, CI [$-$0.170, +0.070] & Equivalent (at $\varepsilon$ = 0.15, stricter than $\delta$ = 0.20) \\
Untrained backbone (same-generation 35B) & +0.240 & --- & --- \\
\bottomrule
\end{tabular}
\end{table*}

\apxsection{E}{Human-Study Instruments}

The three human tasks of \S{}6.2 share one set of 5 raters; the corpus is entirely Chinese, and each rater completes 540--544 items in about 7.5 hours.

\textbf{Study A (the single-turn discrimination task, the main task).} A teacher-forced two-alternative forced-choice discrimination: raters are given the preceding context of a session drawn from the real-conversation set together with two candidate next user utterances --- the one the real person actually said at that moment, and the one the release candidate M3b generates in the same context --- and are asked which came from the real person. Left and right are randomised and the judgement is blind. The main block's 300 contexts are crossed with four candidate sources (the release candidate, two external models, and a human-against-human control), each context $\times$ source cell judged exactly once, so each source nominally receives 300 judgements; there are in addition an overlap block of 80 items (rated by all 5 raters, used to estimate agreement between raters) and 10 attention-check items. The release candidate said nothing at all on 4.65\% of the contexts and those items are excluded, so the reading is written in full as ``on the 95.35\% where it produced an utterance, the discrimination accuracy is 0.622'' --- the 37.8\% reported in \S{}6.2 being its complement; after that exclusion its main block and its share of the overlap block come to 360 judgements.

\textbf{The human-against-human control was the stopping criterion, and the readings triggered it.} The five raters read 0.79 to 0.91 on it, all far from 0.50. This paper did not act on the trigger, having judged that its expected value was never 0.50: both of its options were written by real people and only one of them follows on from the conversation, so topical coherence is itself a usable cue and this arm poses a different and easier question. The cost is recorded as it stands: this study therefore has no control that genuinely sits at chance, and conclusions of the form ``people cannot tell them apart'' lose one layer of protection.

\textbf{Study B (whole-session preference, the secondary task).} The study uses 100 persona-matched pairs. Each pair consists of two multi-turn sessions in which the ``user'' side is played by M3b and by \texttt{gpt-5.3} respectively, all other conditions being equal. The judgement is blind with left and right randomised, each pair is judged once by each of 3 raters, and the question is which of the two sessions is more like a real person. Two LLM judges had given opposite orderings on these pairs in the pilot --- \texttt{deepseek-v4-pro} ranking M3b above \texttt{gpt-5.3}, Claude the reverse --- and a third could not arbitrate; this task is that arbitration. M3b is chosen in 205 of the 300 judgements.

\textbf{The third task (human reproduction of the depth layers).} 145 disclosure items, drawn from 25 personas, are labelled by 5 raters in full crossover, each rater assigning the depth layer of the item in a three-way forced choice; with the original label as the row and the majority vote of the 5 raters as the column, per-layer agreement is \texttt{surface} 77\% $\cdot$ \texttt{mid} 47\% $\cdot$ \texttt{core} 35\%, and Fleiss' $\kappa$ is 0.344 overall, 0.410 on the anchor items (\texttt{observed $\times$ surface}, 53 items), and 0.138 on the contested items (\texttt{inferred $\times$ mid/core}, 51 items; the remaining 41 fall in mixed cells). The design has two features: the original label does not enter the verbatim materials, and no ``uncertain'' option is offered --- the latter puts the guessing baseline of a three-way choice at exactly 33\%, so per-layer agreement can be compared against it directly.

A fourth task (whether the scale agrees with human perception): this paper does not have that evidence. Testing whether the judge's rubric scores agree with human perception needs a task of its own, which this paper does not supply; that fourth question's conclusion may not be carried over from the three tasks above, which answer single-turn discrimination, whole-session preference, and agreement on layer labels --- none of them answers whether the scale tracks human perception.

\apxsection{F}{Provenance of the Corpus and the Gate Labels}

\subsection*{F.1 The Filtering Chain for the Real Corpus and the Stratified Sample}

The chain here is the sampling protocol of the real-person anchor on the evaluation side, not the three exclusions on the training side in \S{}4.2; Table F1 gives the segments remaining at each step.

\begin{table*}[t]
\centering\footnotesize
\setlength{\tabcolsep}{4pt}
\caption{\textbf{Table F1} --- The filtering chain and stratified sample of the real conversation set}
\begin{tabular}{@{}>{\raggedright\arraybackslash}p{0.170\textwidth}>{\raggedright\arraybackslash}p{0.622\textwidth}>{\raggedright\arraybackslash}p{0.139\textwidth}@{}}
\toprule
Step & How it was done & Segments remaining \\
\midrule
Source file & The full set of real conversations from production & 34,919 \\
Basic filtering & Three kinds removed: scenarios outside the scope of evaluation; no disclosure-inventory annotation; acute-crisis scenarios & 30,312 \\
Three predicates specific to this experiment & Heavily occult content (fortune-telling and the like) excluded; one segment kept per person; a disclosure inventory required to hold at least 4 items covering both mid and core & \textbf{18,383} \\
Stratified sampling by conversation quality & 333 drawn from each of the good, mixed, and poor strata (Table F2) & \textbf{999} \\
\bottomrule
\end{tabular}
\end{table*}

One segment per person keeps a single user from being counted repeatedly. The third predicate excludes by construction conversations with no deep material to disclose, so the real-person depth levels here are inherently high and may not be taken as an unbiased estimate of real disclosure depth in production (\S{}6.2, Limitations).

\begin{table*}[t]
\centering\footnotesize
\setlength{\tabcolsep}{4pt}
\caption{\textbf{Table F2} --- The sampling rates and sampling weights of the three strata}
\begin{tabular}{@{}>{\raggedright\arraybackslash}p{0.144\textwidth}>{\raggedright\arraybackslash}p{0.299\textwidth}>{\raggedright\arraybackslash}p{0.103\textwidth}>{\raggedright\arraybackslash}p{0.165\textwidth}>{\raggedright\arraybackslash}p{0.185\textwidth}@{}}
\toprule
Stratum & Segments in the population & Drawn & Sampling rate & Sampling weight \\
\midrule
good & $\approx$ 653 & 333 & 51.0\% & \textbf{1.96} \\
mixed & $\approx$ 15,501 & 333 & 2.1\% & \textbf{46.55} \\
poor & $\approx$ 2,231 & 333 & 14.9\% & \textbf{6.70} \\
Total & $\approx$ \textbf{18,385} (agreeing with the filtered population of 18,383, the difference being rounding) & 999 & 5.4\% & --- \\
\bottomrule
\end{tabular}
\end{table*}

A sampling weight is the reciprocal of the sampling rate --- how many segments of the population each sampled segment stands for. Equal numbers are drawn from strata differing in size by more than twentyfold, so the mixed stratum, the overwhelming majority, is heavily down-sampled and the rare good stratum heavily over-sampled (3.6\% of the population, 33.3\% of the sample). Any mean pooling across strata must state at the point of use whether it is weighted: unweighted describes these 999 segments, and only weighting extrapolates to the 18,383.

\subsection*{F.2 The Per-Scenario Rating Distribution of the Real Corpus}

\S{}4.1 gives only the two extreme readings; the whole-corpus distribution and the per-scenario extremes follow. The denominator is always the rated conversations in that scenario (the per-scenario split of the 31,197 rated conversations --- a superset of the 30,312 of Table F1, since rating precedes basic filtering); the proportion is the share rated good on the companion agent's overall performance.

Over the whole set: 712 rated good (2.28\%), 19,623 mixed (62.9\%), 10,862 poor (34.8\%). The two high-risk extremes of \S{}4.1 are gender scripts at 0.00\% (0/28) and acute crisis at 0.12\% (1/868); the three low-risk controls fall between 10.66\% and 15.74\%, an order of magnitude above them; at cell level the same shape holds --- of the 27 cells below the sample floor, 19 (70\%) contain no anchor rated good at all.

\textbf{The scarcity of pairs behind \S{}4.1's second gap is also read here.} Of 6,734 rated users only 147 (2.18\%) have both a conversation rated good and one rated poor; restricting to those with more than one conversation (3,368) still gives 4.36\%. The gate's earning side is a 2\% minority class in the real corpus to begin with, and close to unobservable exactly where the gate should matter most. The readings above are the complete evidence for \S{}4.1's first gap, and this paragraph for the second; together they are why the seed pool below retains every seed rated good without exception.

\textbf{The seed pool and the filling of cells.} The real annotations also serve as seed personas. After exclusion and de-duplication the pool holds about 24,000 seeds over 6,500 users, with the companion agent rated good in only 2.4\% of them; filtering takes 10,000 seeds, every seed rated good retained. The filling prompt supplies at most 4 real personas as examples, from which entirely new personas are generated following their layering and defensiveness: content new, form inherited. Seeds are arranged into 68 cells with a floor of 50; 27 cells fall below it and 3 cells hold no real sample, so about 900 synthetic seeds are generated to fill them --- of which nearly 200 require expert review and do not enter the training set; for about 90 of those, no trajectories are generated at all.

\subsection*{F.3 Annotating the Inventory on the Real Branch}

\textbf{One call, taking the whole conversation and its existing structured annotation (including the production-side profile).} The annotator \texttt{claude-sonnet-4-6} breaks the user's inner material into typically 3--6 items, each given four labels (layer, gate, guard, may-never-surface) and two provenance fields (source \texttt{observed} / \texttt{inferred}; confidence 1--5).

Two hard constraints are relied on. Coverage: the inventory must span surface through core and contain at least one surface item whose content the opening line is to say --- otherwise there is nothing to say in the first turn, the simulator invents an item not in the inventory, and the depth readings lose their anchor. Wording: content is always a third-person summary; staying close to the original words is forbidden, since close wording is recited verbatim by the simulator and carries a real user's phrasing into the model's weights.

That first constraint and the synthetic side's ``the inventory is fixed before the opening line'' (\S{}4.2) are two sides of one requirement: on the real side a constraint makes inventory and opening line match, on the synthetic side the emission order does it.

\textbf{A per-segment turn cap and character cap are set and two cleaning passes applied} --- hygiene, not affecting label semantics.

\textbf{The rules governing the two annotation conflicts.} Premature disclosure --- the gate is not satisfied yet the user has already spoken the item --- is resolved as follows: if the item would be spoken even to a cold or neutral AI, it was lightly defended to begin with and its gate is lowered; if the disclosure came from a one-off external force while the item is defended in essence, the high gate is kept and the item is flagged as a disposition-divergent premature disclosure. Where a gate is open while the item stays covered, the gate is left unchanged by default, disclosure being permitted rather than compelled; it is raised, or the item relabelled as one that may never surface, only where the user repeatedly meets responses that can hold the material and it still never surfaces.

\textbf{The three exclusions on the training side.} Every conversation involving a persona used in evaluation is removed at the level of the user (otherwise the simulator might recognise at evaluation a persona seen in training); so are conversations outside emotional companionship, and acute-crisis scenarios. Of what remains, the 4 conversations with the most trainable turns are taken per user --- a selection rather than a truncation, biased towards conversations with more turns. The three exclusions leave the training real branch at about 13,000 conversations and 286,000 trainable user turns (\S{}4.2).

\subsection*{F.4 Depth-Label Confidence by Layer, and Its Bearing}

The annotator writes the inventory out item by item after reading the whole conversation and the profile (procedure in F.3), recording for each its depth layer, whether it was observed directly (\texttt{observed}) or inferred by the annotator (\texttt{inferred}), and confidence 1--5.

\textbf{Inferred} means the item was never spoken but was reasoned out from the conversation and the profile on the user's behalf. It is necessary because a poorly performing companion agent never earns the core layer, so items there never get the chance to be spoken, and admitting only the directly observed would leave not a single deep item --- whereas what the simulator needs is exactly an inventory of ``what the user has not yet said and could be earned''.

\textbf{Source and confidence both vary systematically in the same direction along the depth axis}, and two persona corpora show the same shape (Table F3):

\begin{table*}[t]
\centering\footnotesize
\setlength{\tabcolsep}{4pt}
\caption{\textbf{Table F3} --- Corpus $\times$ depth $\times$ source $\times$ confidence}
\begin{tabular}{@{}>{\raggedright\arraybackslash}p{0.228\textwidth}>{\raggedright\arraybackslash}p{0.121\textwidth}>{\raggedright\arraybackslash}p{0.076\textwidth}>{\raggedright\arraybackslash}p{0.152\textwidth}>{\raggedright\arraybackslash}p{0.152\textwidth}>{\raggedright\arraybackslash}p{0.152\textwidth}@{}}
\toprule
Corpus & Depth layer & \textit{n} & \% \texttt{inferred} & Mean confidence (1--5) & \% confidence $\leq$ 3 \\
\midrule
The intrinsic-layer persona set (ZH) & surface & 209 & 0.0\% & \textbf{4.93} & 0.0\% \\
 & mid & 262 & 7.3\% & 4.08 & 9.2\% \\
 & \textbf{core} & 123 & \textbf{88.6\%} & \textbf{2.37} & \textbf{94.3\%} \\
The larger Chinese persona corpus (500 personas) & surface & 1,019 & 0.0\% & \textbf{4.94} & 0.1\% \\
 & mid & 1,396 & 7.4\% & 3.97 & 17.4\% \\
 & \textbf{core} & 758 & \textbf{58.4\%} & \textbf{2.34} & \textbf{96.0\%} \\
\bottomrule
\end{tabular}
\end{table*}

The semantics promised by the deepest of the three depth layers specified in \S{}3 is therefore stronger than the data can support (Limitations).

One inseparability must be recorded alongside: ``remove only the deepest layer'' and ``remove only the low-confidence labels'' differ by $\leq$ 0.03 and order the candidates identically --- 94\% to 96\% of the core layer coincides with the low-confidence layer, so on this data they are almost one operation. Whether it is a problem of the layer or of label confidence therefore cannot be answered on the data available; since the core layer's semantics already fails to reproduce (the per-layer agreement rates in Appendix E), adding annotators does not resolve it, and the Limitations section puts the follow-up on a depth measure not depending on labels at the deep end.

\textbf{The robustness check ``keep only the high-confidence labels'' must be reported for what it is}: by layer it amounts to looking at surface and mid only. Under it the condition-permutation test remains significant for all 11 candidates.

\textbf{The conclusion does not depend on the core layer.} The agreement of human reproduction for that layer's labels is in Appendix E.

\subsection*{F.5 Which Contract Fields Enter the System Prompt}

F.3 gives the annotation procedure; this section gives which of the several dozen fields of the annotation contract enter the simulator's system prompt once annotation is complete --- the renderer shared by the training and inference ends takes only some (the complementary half is Appendix J's stripping, which states what is removed on top of this). What is taken: the standing context; the invariants; the summary of previous conversations (empty for a first session); the disclosure inventory with the gate legend compiled from it; and the anti-goals (per-role visibility in Appendix B) --- plus the opening line (folded in so the companion agent opens the conversation); the seven items of the profile (age, gender, attachment type, current emotion, whether the user tends to minimise, tone of voice, what the user wants); and the six of speech style (turn granularity, message length, punctuation, register, surface features, stock phrases). Everything else is discarded, in three classes.

\textbf{Factual} fields --- occupation, city, interests, personality inventories and so on, more than twenty items --- are already carried by the narrative of the standing context, and re-listing them would only conflict with it. Inner-state fields --- Kohutian needs, self-state, secondary emotion, emotional intensity, active defences, cognitive distortions, developmental stage, crisis level, existential concerns, transference object --- are left to the fine-tuning to internalise rather than declared in the prompt, by the same reasoning as B.4's ``compile what can be devolved, internalise what cannot''. Metadata of the disclosure inventory --- source (\texttt{observed} / \texttt{inferred}), confidence (1--5), and the disposition-divergence flag (\texttt{realized\_divergence}) --- is discarded too: the first two are annotation metadata not needed for the role-play, while discarding the third is a real gap, since that field was introduced to down-weight the corresponding turns in training, and not carrying it over means the down-weighting has not been applied (Limitations).

Training and evaluation share one rendering function --- the real branch's construction script imports the synthetic branch's renderer directly --- so a change to the field selection is made in one place only (\S{}4.3).

\subsection*{F.6 The English Transcreation: Rules, Checks, Boundaries}

\S{}4.2 keeps only the conclusion; the reason, the adaptation rules, the acceptance checks, and the validity boundaries are gathered here.

\textbf{The English corpus is a transcreation of the Chinese trajectories --- not an independent English re-run and not a literal translation.} Each Chinese trajectory yields its English version in one call with the shape kept exactly: N turns in, N turns out, the same speaker in the same position, the same items disclosed and withheld per turn; turn alignment is a hard constraint, so the per-turn move annotations are re-attached by position unchanged. The personas first undergo one offline adaptation in three layers: structural and enumerated fields are kept as they are (change one and the per-turn move labels become invalid); narrative fields are translated and culturally adapted; the opening line and tone of voice are regenerated natively. This produces 10,780 English seeds.

\textbf{The first reason is not cost but that it produces a strictly paired Chinese--English corpus.} Each English trajectory shares with its Chinese original one persona, one disclosure inventory, one string of per-turn move annotations, and one state-machine trajectory, differing only in surface language and its cultural adaptation. A cross-language comparison can therefore be made item by item on one batch of trajectories rather than between two independently drawn samples: the disclosure inventory, the move annotations, the scenario distribution, and the sequence in which the gates open and close are identical by construction, so the remaining difference can only be attributed to language and to the models trained on it. Two independently synthesised sides share no underlying trajectory, and any cross-language difference would be confounded with the difference between the corpora themselves.

\textbf{Cost is the second reason.} The multiple varies with the protocol and must be cited with it: transcreation makes one call per whole conversation, whereas \S{}4.2's trajectory generation advances turn by turn, one generation per role per turn plus retries. The three protocols give about 14$\times$ by the amount actually paid (the trajectory counts being almost identical), about 4$\times$ by the unit price of one synchronous call, and about 10$\times$ by that of one official batch call. Citing ``about fourfold'' alone cites the narrowest of the three.

\textbf{Linguistic quality was settled by one further blind assessment.} A judge model not involved in generation decided whether the text read as native: both transcreated examples passed (5.0 / 5.0), both literal translations used as a discriminating control failed (3.0 / 2.0). The gap between transcreation and a native re-run is therefore smaller than that between transcreation and literal translation --- the transcreated text sits at the ceiling of the nativeness scale --- but the check covers linguistic naturalness only.

\textbf{Five zero-tolerance acceptance checks}: an illegal structure; a misplaced turn or speaker; residual Chinese characters; a truncated ending; the output cap being reached. Where a check fails but English was produced, the trajectory is recovered after human review --- the checks flag rather than discard; 22,363 of 22,365 unique trajectories were admitted.

\subsection*{F.7 The Corpora}

Table F4 collects the corpora this paper uses.

\begin{table*}[t]
\centering\footnotesize
\setlength{\tabcolsep}{4pt}
\caption{\textbf{Table F4} --- The corpora of this paper (evaluation and training; the re-run of the real-conversation set has a row of its own)}
\begin{tabular}{@{}>{\raggedright\arraybackslash}p{0.230\textwidth}>{\raggedright\arraybackslash}p{0.108\textwidth}>{\raggedright\arraybackslash}p{0.360\textwidth}>{\raggedright\arraybackslash}p{0.215\textwidth}@{}}
\toprule
Corpus & Language & Size & Used for \\
\midrule
Intrinsic-layer persona set & ZH / EN & 100 personas each $\times$ 3 conditions $\times$ 20 turns & \S{}6.1, \S{}6.3 \\
Extension persona set & EN & 321 wholly new personas, disjoint from the intrinsic-layer set & Independent replication (\S{}6.3) \\
Real-conversation set & ZH & 999 production sessions (333 each of good / mixed / poor) & Human anchor (\S{}6.2) \\
Re-run under a consistent protocol & ZH & 600 sessions (the simulator re-run on the real set's personas and rendering) & Making the two sides comparable (\S{}6.2) \\
Real held-out set & ZH & 500 sessions / 12,437 real turns & The primary reference for distributional fidelity, and the unseen-user split for perplexity (\S{}6.2, \S{}6.3) \\
Human-study verbatim materials & ZH & Three tasks; item counts in Appendix E & Human discrimination, preference, and layer annotation (\S{}6.2) \\
Downstream corpus & EN & 12 SUTs $\times$ 100 personas $\times$ 7 evaluation environments = 8,400 sessions of 20 turns & \S{}6.4 \\
Chinese training corpus & ZH & $\approx$ 33,000 conversations / 714,000 trainable user turns (synthetic : real 60 : 40) & Training (\S{}4.3) \\
English training corpus & EN & $\approx$ 22,000 conversations / 454,000 turns & Training (\S{}4.3) \\
\bottomrule
\end{tabular}
\end{table*}

\apxsection{G}{A Domain-General Template and Three Instantiations}

\subsection*{G.1 The Four Components and the Properties Each Must Satisfy}

\S{}3 sets this mechanism out as a specification for the companionship domain: five gates, three depths, eight behaviour classes, and seven guards. Strip away these domain-specific names and four components remain --- an ordered state ladder, a transition predicate, asymmetric retreat, and a per-example gate legend (the ladder and the transition predicate in \S{}3.2, retreat in \S{}3.3, the legend in \S{}3.1; \S{}3.4 lists their names).

Why these four? The first three govern one indispensable aspect each --- where the information is held, what earns its release, what a misstep costs --- while the fourth delivers to each example, auditably and at zero marginal cost, the meaning of only those values the example uses, the rest being left to training. None may be omitted: without a ladder there is no depth; without a transition predicate, depth bears no relation to what the other party does; without retreat, an incorrect action costs nothing; without a per-example legend, the model must guess the specification.

What porting to another domain actually replaces is only the names of the tiers and the classes. The difficulty lies in the third column of Table G1 --- the properties each component must carry with it, and they are the easiest thing to lose when the names change.

\begin{table*}[t]
\centering\footnotesize
\setlength{\tabcolsep}{4pt}
\caption{\textbf{Table G1} --- The four components, and the properties each must satisfy when ported to another domain}
\begin{tabular}{@{}>{\raggedright\arraybackslash}p{0.222\textwidth}>{\raggedright\arraybackslash}p{0.325\textwidth}>{\raggedright\arraybackslash}p{0.384\textwidth}@{}}
\toprule
Component & Definition & Must also hold after porting \\
\midrule
\textbf{Ordered state ladder} & An ordered set of states, \textbf{invisible} to the party under test, projected onto a coarse-grained observable layer & Ordered; \textbf{more tiers than there are observable layers} --- otherwise the distinction ``already one step further within the same layer'' is lost (B.1 gives the two instances from the companionship domain) \\
\textbf{Transition predicate} & State transitions are defined by \textbf{the class of behaviour shown by the other party, the party under test}, not by turn index, schedule, or script & \textbf{No gate skipping} (the conditional \texttt{max}, property (i) of \S{}3.2); the classes must be annotatable by a third party --- otherwise the specification is not executable (B.2) \\
\textbf{Asymmetric retreat} & A set of behaviour classes lowers the state \textbf{and sets a rupture flag}, and that flag triggers a guard observable at the surface & Retreat is quick and rebuilding slow; the cost of rebuilding should \textbf{emerge} from the orderedness of the ladder rather than be added as a penalty term (\S{}3.3, B.3) \\
\textbf{Per-example gate legend} & Compiles into the per-example condition, deterministically, only the gate definitions and guard definitions that the example actually uses; the global mechanism is left to training to internalise & Deterministic, zero marginal cost, minimal exposure, and changing a rule does not require rewriting the data (\S{}3.1, B.4) \\
\bottomrule
\end{tabular}
\end{table*}

\subsection*{G.2 Three Instantiations}

Three domains are each given one instantiation, showing what values the four components take when ported; all are design sketches, none implemented or validated, portability being a design claim rather than an experimental result. The transition predicates are examples, not exhaustive.

Each ladder, in outline: in the job interview the candidate's disclosure of true motives --- surface interest in the post $\rightarrow$ the real reason for leaving $\rightarrow$ salary floor and offers in hand $\rightarrow$ personal constraints; in medical consultation the patient's symptom disclosure --- presenting complaint $\rightarrow$ relevant history $\rightarrow$ information that is shameful or stigmatised $\rightarrow$ the concern never voiced aloud; in negotiation the disclosure of one's hand --- position $\rightarrow$ interest $\rightarrow$ constraint $\rightarrow$ the BATNA. Each ladder merges its middle two tiers, giving three observable depth layers --- surface interest / motive and leverage / personal constraints; presenting complaint / history and stigma / the unvoiced concern; position / interest and constraint / the BATNA --- so Table G1's ``more tiers than layers'' holds in all three. The advancing predicates are, respectively, asking behavioural rather than hypothetical questions, restating technical detail accurately, and naming a team difficulty first; asking open rather than closed questions, normalising the symptom, and explaining why one is asking; and probing interest rather than price, and disclosing one of one's own constraints first. Retreat is triggered by passing judgement on a résumé gap, by moralising about poor adherence, and by threatening --- after which the candidate gives standardised answers, the patient turns to \texttt{minimized} (``I take them on time, mostly''), and the counterparty falls back to positional bargaining. In each domain the gate legend includes only the concerns, stigma entries, or constraints the party holding the information actually has.

\subsection*{G.3 The Three Instantiations Share One Shortcut}

The three domains are unrelated, yet the shortcut open to the party under test is the same: scoring through question coverage --- every slot, every symptom, every clause --- rather than through making the holder of the information willing to hand it over. As far as we are aware, existing benchmarks in all three reward exactly that. Interview benchmarks treat the candidate as a static respondent and so cannot separate an interviewer who asks comprehensively from one who makes people willing to speak truthfully. USMLE-style consultation evaluations assume the patient answers honestly, so they cannot measure whether the consulting party created the conditions for disclosure. Negotiation benchmarks mostly use a fixed payoff matrix with a fully rational or scripted counterparty, so they cannot measure whether the AI turns a distributive negotiation into an integrative one. What all three reward is exhausting the questions, whereas what this template sets out to reward is something else: making the party holding the information willing to answer. Each component forecloses one segment of the shortcut: the ladder makes the information impossible to obtain all at once; the transition predicate makes obtaining it depend on the quality of the behaviour rather than its quantity; asymmetric retreat attaches a cost to incorrect behaviour; and the per-example gate legend makes the other three take effect in every example, at zero marginal cost, rather than existing only in the specification.

\subsection*{G.4 Conditions of Applicability, and Where the Template Does Not Apply}

The template requires a structure in which the flow of information is determined by the other party's behaviour and the behaviour classes can be annotated by a third party (otherwise the transition predicate is not executable). Two kinds of case fall outside it. Purely task-oriented dialogue (booking a flight, checking a balance) has no latent ladder: the asymmetry is a matter of task rather than relationship, slot filling is the better model, and forcing the template on it imposes no constraint --- the score still reflects nothing but question coverage, precisely the shortcut it was meant to foreclose. Evaluations in which the party holding the information has no motive to conceal it (the examiner in a question-answering benchmark) leave asymmetric retreat nothing to attach to: with no relationship that can be damaged, retreat cannot be defined, and the template degenerates into a source that hands the information over as soon as it is asked for.

\apxsection{H}{Twelve Rejected Designs}

This appendix advances no new claim; it gives the reasons behind the specification. The twelve are grouped by the nature of the reason for rejection, and the table is ordered accordingly. Principle (6, \#1--\#6): adopting one of these would destroy a property this paper depends on --- discriminating power, auditability, control over the state space rather than the path, or observability of the learning signal. Cost (3, \#7--\#9): the reason is budget or engineering effort, not principle, so any whose budget constraint still binds should be reopened when resources allow, with the cost of the concession stated. Delivery (3, \#10--\#12): adopting one of these would not change the semantics of the mechanism, only how it is implemented or issued.

Within the principle class, \#2, \#4, and \#6 share one failure mode and are grouped accordingly in the closing paragraph; the remaining six of the twelve (\#7--\#12) are the cost and delivery classes. These twelve are not an exhaustive account of the design space; they are the alternatives this paper actually considered and recorded. Table H1 gives them one by one.

\begin{table*}[t]
\centering\footnotesize
\setlength{\tabcolsep}{4pt}
\caption{\textbf{Table H1} --- The twelve rejected designs (rejected design $\cdot$ why it is attractive $\cdot$ why it was rejected $\cdot$ when to reopen; the class of each is given above)}
\begin{tabular}{@{}>{\raggedright\arraybackslash}p{0.024\textwidth}>{\raggedright\arraybackslash}p{0.196\textwidth}>{\raggedright\arraybackslash}p{0.179\textwidth}>{\raggedright\arraybackslash}p{0.301\textwidth}>{\raggedright\arraybackslash}p{0.197\textwidth}@{}}
\toprule
\# & Rejected design & Why it is attractive & Why it was rejected & When to reopen \\
\midrule
1 & \textbf{Disclosure on a schedule}: fixing in advance how deeply each turn discloses & Maximally controllable and fully reproducible; needs neither annotation nor a judge & Discriminating power is zero by construction: good and trips receive the same material on the same turn, and the environment produces no information about the party under test & Never (the gate replaced it) \\
2 & A hand-written \textbf{per-session transition policy} & Behaviour is explicitly controllable and can be fitted segment by segment to real trajectories & Where it conflicts with the real trajectory the model has no basis to choose, and it delivers the path rather than the state space & Never (superseded by ``disposition as policy'': the disposition field is the condition, transitions turn on the other party's behaviour alone) \\
3 & Implementing the gate as a \textbf{learned classifier} & Data-driven, coverage extends automatically, and no hand-written decision criteria are needed & Not auditable: no clause can be faulted point by point, only wholesale acceptance or rejection, and changing a rule requires rewriting the data & When behaviour outside the eight \texttt{ai\_move} classes becomes common --- that is, when rule coverage is insufficient \\
4 & Issuing the simulator a \textbf{whole-session interaction-style profile} & Supplies information about the user's interaction style & A post-hoc profile summarises the session and contains its outcome, so the simulator converges on a known endpoint (the path is controlled); it was removed from the judge side for the same reason & Never (containing the outcome controls the path, whatever the profile's quality) \\
5 & Inferring the gate and the counterfactual from \textbf{sessions in which the companion performed poorly} & Uses all available data (such sessions are the overwhelming majority) & Unobservable in principle: unlocking never occurs there, and ``what if it had empathised'' leaves no trace in the data & Never (replaced by slices of good sessions plus controlled synthesis) \\
6 & Issuing the simulator a \textbf{complete event line} at runtime & Supplies a skeleton script and improves continuity & Redundant with the inventory, and a whole event line contains the outcome, so the path is controlled & Not reopened --- downgraded to factual grounding at synthesis, not issued at runtime \\
7 & \textbf{Per-turn ground-truth annotation} (AI behaviour quality, realised disclosure order, and emotional trajectory on every turn) & A fine-grained supervision signal; cross-turn conditions could then be enforced from data rather than at the annotator's discretion & Cost: per-turn annotation of thirty-five thousand sessions is not feasible within this paper's budget & When the budget allows; minimally, a small annotation sample checking that the annotator's enforcement of the continuity condition agrees with the state machine's orderedness \\
8 & A \textbf{rolling cross-session inventory} (striking out what was disclosed last time, adding what newly emerged) & A more complete seed for continued multi-session conversation & The hardest of all the fields to roll forward: it must both remove what was disclosed and generate what newly emerged, each introducing uncontrollable drift & When multi-session continuation is deepened \\
9 & Generating all twenty-two thousand sessions by \textbf{real-time concurrency} & Results are visible immediately and the pipeline is simpler & Neither the time nor the cost is acceptable & Not reopened --- acceptable once the batch interface was adopted \\
10 & An \textbf{initial-stance} field (how defended the user is at the start) & Fixes the starting point and reduces first-turn variance & Redundant: the opening utterance already implies the initial stance & If the opening utterance proves not to carry the starting point \\
11 & Having a continued session's opening \textbf{reuse the closing fields} of the previous session & Saves one generation call & Inconsistent once the history is reintegrated; regenerating from the new history was adopted & --- \\
12 & Fixing the \textbf{guard wording of every item} in advance & The linguistic realisation of the guard is more vivid & Redundant: generated at runtime from the persona's register, saving an annotation pass & If the seven generic guard descriptions prove insufficiently vivid \\
\bottomrule
\end{tabular}
\end{table*}

An instrument fit to serve as an evaluation environment must produce information about the party under test, be open to being faulted point by point by a third party, control the state space rather than the path, and leave the learning signal traceable on the surface of the text; \#1, \#3, \#2 / \#4 / \#6, and \#5 each fail one of the four, and those four are the failure modes this appendix is organised by. Only when all four hold at once can the shifts in the downstream ranking and readings that follow from replacing the simulator be attributed to the environment itself (the two acceptance criteria of \S{}6.4 rest on this).

\clearpage

\apxsection{I}{The Model Variants}

Table I1 defines, row by row, the model labels used throughout the paper. The trained variants are fine-tuned from two mixture-of-experts backbones, \texttt{Qwen3.5-35B-A3B} at 35B and \texttt{Qwen3.5-122B-A10B} at 122B; the Chinese variants train for 2 epochs and the English for 3, with every other optimisation setting identical across variants in one language, so that any two variants within one language differ in one variable only. Every rollout in the paper decodes at temperature 0.7 --- also the temperature used to produce the published leaderboard this paper compares against --- and the release defaults are set per language: T = 1.0 for Chinese, T = 0.7 for English, with a shared hard ceiling of T $\leq$ 1.0 (the Chinese default of 1.0 rests on a temperature sweep not reported here). The companion agent is part of the environment protocol rather than an object of measurement: the good block is fixed to \texttt{gpt-4.1} and the neutral and trips blocks to \texttt{gpt-4.1-mini} (the weaker model performs the perfunctory and anti-goal specifications more readily), the trips block triggering the anti-goal behaviour by specification; all three are held constant across every intrinsic-layer comparison.

\begin{table*}[t]
\centering\footnotesize
\setlength{\tabcolsep}{4pt}
\caption{\textbf{Table I1} --- The model variants (the sole definition of the labels used throughout)}
\begin{tabular}{@{}>{\raggedright\arraybackslash}p{0.189\textwidth}>{\raggedright\arraybackslash}p{0.223\textwidth}>{\raggedright\arraybackslash}p{0.111\textwidth}>{\raggedright\arraybackslash}p{0.111\textwidth}>{\raggedright\arraybackslash}p{0.123\textwidth}>{\raggedright\arraybackslash}p{0.123\textwidth}@{}}
\toprule
Label & Scale & Language & Training branch & Gate information at training time & Gate information at inference time \\
\midrule
M1 / M1b $\dagger$ & 35B / 122B & EN & synth-only & Intact & Intact \\
M2 / M2b & 35B / 122B & EN & synth-only & Stripped & Stripped \\
M3 / M3b $\dagger$ & 35B / 122B & ZH & synth + real & Intact & Intact \\
M4 & 35B & ZH & real-only & Intact & Intact \\
M5 & 35B & ZH & synth-only & Intact & Intact \\
A1 / A2 / A3 & M1 weights & EN & --- & Intact & Intact / Stripped / Mis-mounted \\
Reference simulator (EN) & 235B (previous-generation backbone) & EN & synth-only & Intact & Intact \\
Reference simulator (ZH) & 235B (previous-generation backbone) & ZH & synth + real & Intact & Intact \\
Untrained backbone & 35B / 122B / previous-generation 235B & EN (235B bilingual) & --- & --- & Intact (35B and 122B also tested stripped) \\
\texttt{gpt-5.3} & --- & Bilingual & --- & --- & Intact / Stripped (two states) \\
\texttt{deepseek-v4-pro} (also the main judge) & --- & Bilingual & --- & --- & Not applicable \\
\texttt{gpt-4.1} & --- & Bilingual & --- & --- & Not applicable \\
\texttt{gpt-4.1-mini} & --- & Bilingual & --- & --- & Not applicable \\
\texttt{gpt-5.1} & --- & EN & --- & --- & Not applicable \\
\bottomrule
\end{tabular}
\end{table*}

$\dagger$ marks the 122B release candidates M1b and M3b; each of the first three rows covers two scales. The release artifacts and data provenance are in Appendix K.

Each external model has a role in the paper: \texttt{gpt-5.3} appears intact and stripped, testing whether the gating transfers without training (Appendix D); \texttt{deepseek-v4-pro} is the external reference for distributional fidelity and the main judge of \S{}5.4; \texttt{gpt-5.1} serves as a downstream evaluation environment in \S{}6.4. The two 235B reference simulators produced the published leaderboard compared against here \citep{companionbench} and are the baseline for downstream displacement, and the untrained backbones --- the same pretrained models the trained variants were fine-tuned from --- are the floor known in advance.

\clearpage

\apxsection{J}{Field-Level Definitions of the Two Operations on the Gate Information}

This appendix gives the verifiable definitions of the two operations on the gate information, defined in \S{}5.2: stripping removes gate information, the placebo supplies the wrong mounting relations while leaving the shape identical. The rule set is in Appendix B. Fig J1 sets the two side by side, field by field. Fig 2 (\S{}4) places the four manipulations of \S{}1 --- the two gate-information operations among them --- on the pipeline they cut.

\begin{figure*}[t]
\centering
\includegraphics[width=1.00\textwidth]{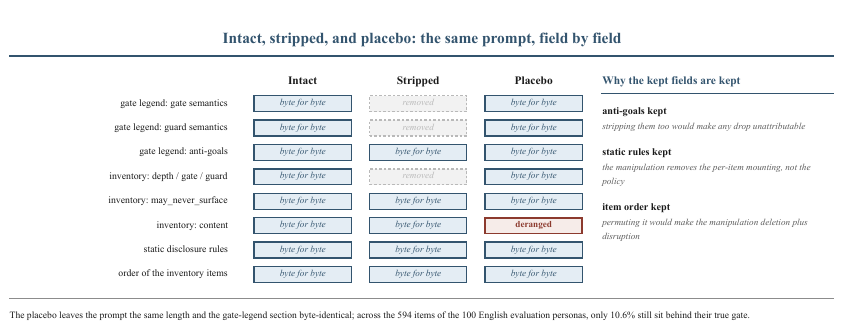}
\caption{\textbf{Fig J1} --- Intact, stripped, and placebo: the same prompt, field by field (stripping deletes the two semantics sections and three annotation fields; the placebo deletes nothing and only deranges \texttt{content} within a persona)}
\end{figure*}

\subsection*{J.1 Stripping}

Stripping is implemented by a single function shared by the training and inference paths: at training time it rewrites the corpus into the M2 / M2b datasets, at inference time it is imported through an in-process wrapper replacing the renderer --- so what is removed at the two time points is the same object.

\textbf{Two elements are removed.} The first is the leading two sections of the gate legend --- the gate semantics and the guard semantics (\S{}3.1); the third section, the anti-goals, is retained, as set out below. The second is the \texttt{depth}, \texttt{gate}, and \texttt{guard} fields of every item in the disclosure inventory, leaving only \texttt{content} and \texttt{may\_never\_surface}; the depth claim in the inventory heading is removed with them, since \texttt{depth} no longer exists.

\textbf{Four elements are retained.} Each item's \texttt{content} and \texttt{may\_never\_surface} (above). The anti-goals (\texttt{\# Your landmines (anti\_goal: $\ldots$)}) --- the persona's own retreat triggers rather than gate semantics; stripping them too would make any decrease in the trips condition unattributable between ``the gate information was load-bearing'' and ``the simulator was never told this persona's anti-goals''. The static template of the disclosure policy (\texttt{\# The disclosure rules (internalize these)}), which states the policy --- gates are ordered, must be earned one at a time, default is to withhold --- and is word-for-word identical in every record: what this manipulation removes is the per-item mounting, not the policy. And the order of the inventory items, since permuting it would turn the manipulation into ``deletion plus disruption'', under which any degradation would be unattributable.

\textbf{The boundary of the claim.} This variant supports only the claim that the simulator is no longer told which disclosure sits behind which gate, and how to cover it while the gate is closed --- not that it ``can no longer tell which disclosure is deeper'': across the 21,645 records of the stripped corpus (the full English corpus of F.6 holds 22,365 trajectories), 95.07\% of inventories are already sorted surface $\leq$ mid $\leq$ core and 99.39\% of \texttt{may\_never\_surface} items are core, so with \texttt{depth} gone the depth ordering is still readable from position alone.

\subsection*{J.2 The Placebo}

\textbf{Why this variant exists.} Every SFT training example carries gate information, so deleting it at inference time (A2) presents an out-of-distribution prompt, and a decrease could be attributed to ``this prompt shape was never seen in training'' rather than ``the gate information was load-bearing''. M2 / M2b are retrained on stripped corpora, for which the shape is in distribution, so they are not subject to this threat. A placebo must keep the shape unchanged and alter only the information.

\textbf{What it does.} The annotation columns (\texttt{depth}, \texttt{gate}, \texttt{guard}, \texttt{may\_never\_surface}) stay byte for byte in their original slots, and only the \texttt{content} strings are deranged within a persona; the permutation leaves the total length of the prompt unchanged. The prompt therefore still says ``item 5 is \texttt{core}, unlocked at \texttt{earned\_deep\_trust}, covered by \texttt{rationalization}, and may never surface'', except that this description is now attached to the wrong disclosure; across the 594 items of the 100 English evaluation personas, only 10.6\% of the items still sit behind their true gate, while the gate legend section is byte-identical and the remaining structural markers show no measurable difference from those of a genuine prompt. This is the exact inverse of stripping: stripping withholds which disclosure sits behind which gate, whereas the placebo asserts it wrongly.

\textbf{Why the derangement is applied to the material, not the annotation.} Three alternatives were rejected, each failing in a different way. Swapping the annotation triples across personas returns most of the legend to the model (66.3\% of items keep their true gate) and leaves \texttt{may\_never\_surface} on items relabelled shallow. Swapping only \texttt{gate} and \texttt{guard} within a persona, leaving \texttt{depth} truthful, produces impossible (\texttt{depth}, \texttt{gate}) combinations for 64.8\% of items. Swapping the whole annotation group within a persona yields legal combinations but disrupts the depth ordering (monotone inventories fall from 97/100 to 0/100), so the prompt is visibly not real. Each either returns the information or adds a detectable trace, so none qualifies as a placebo.

\textbf{One irreducible residual cost.} The mispairing is semantic: a plainly deep item (``a fear of not being worth loving at all'') may sit in a slot annotated shallow --- undetectable by any field-level check, and potentially noticeable as an incongruity to a capable model. Any placebo must mispair content with gate information somewhere; this design confines it to the place the experiment intends.

\apxsection{K}{Release Artifacts and Data Provenance}

\textbf{Two 122B simulators are released}: M1b on the English side, trained on the synthetic branch alone, and M3b on the Chinese side, trained on the synthetic and real branches together; their labels are in \S{}5.2 and Appendix I, and the release is at the 122B scale rather than the 35B scale of the same recipe, the reason being reproducibility of the published leaderboard rather than gating strength (\S{}7). Both models are released with a model card stating the default decoding settings (Appendix I) and the two models' asymmetry in downstream evidence: the seven environments of \S{}6.4 are all on the English side, so M1b's two acceptance criteria --- order-preserving and scale-stable --- were both measured, and both are met, while M3b has downstream evidence for neither (Limitations).

Both simulators and all accompanying artifacts are prepared for public release upon publication. \textbf{The non-model artifacts} each come with a run manifest whose fields follow the same rule as the runs reported in this paper: dataset version, configuration hash, schema version, and random seed. They are: the full gating specification (the complete version of Appendix B, not previously public); the audit toolchain and every judge prompt (Appendix C in executable form); the permutation reference-distribution code (which produces the reference distributions of Appendix D); synthetic trajectories only --- 904 synthetic seeds together with the 2,433 trajectories they yielded (not every seed yielded one), neither containing any real user content; the human-study verbatim materials and their instructions (Appendix E); 500 paired bilingual benchmark items, the release promised by the published version; and code for rollout, audit, statistics, and figures.

\textbf{The real corpus consists of de-identified, licensed production logs}, and its use and redistribution rest on three rulings: one restriction, one permission, and one avoidance. The restriction is that the text of real conversations is visible to internal annotators only and enters no release artifact; the training trajectories released are synthetic and contain no real user content. The permission is that aggregate behavioural statistics over real conversations may be published --- the real-side readings of \S{}6.2 stand on that. The avoidance is that the 235B weights are not released --- the two reference models registered in \S{}5.2. Withholding them avoids the question of whether the corpus licence covers redistribution of weights, and it reflects that the smaller model is the artifact others can actually run.

\end{document}